\documentclass[twoside]{article}

\usepackage{PRIMEarxiv}

\usepackage[utf8]{inputenc} 
\usepackage[T1]{fontenc}    
\usepackage{hyperref}       
\usepackage{url}            
\usepackage{booktabs}       
\usepackage{amsfonts}       
\usepackage{nicefrac}       
\usepackage{microtype}      
\usepackage{lipsum}
\usepackage{fancyhdr}       
\usepackage{graphicx}       
\graphicspath{{media/}}     

\usepackage{moreverb,url}

\usepackage{amssymb,amsmath}
\usepackage{caption}
\usepackage{subcaption}
\usepackage{mathrsfs}
\newtheorem{hypothesis}{Hypothesis}
\usepackage{algorithm, algpseudocode}
\usepackage{natbib}
\def\mathbi#1{\textbf{\em #1}}

\usepackage{enumerate}
\usepackage{todonotes}

\title{Hierarchical Planning and \\Policy Shaping Shared Autonomy \\ for Articulated Robots
}

\author{
  Ehsan Yousefi \\
  Dept. of Mechanical Engineering \\
  McGill University \\
  Montreal, QC, Canada\\
  \texttt{ehsan.yousefi@mail.mcgill.ca} \\
   \And
  Mo Chen \\
  School of Computing Science \\
  Simon Fraser University (SFU) \\
  Burnaby, BC, Canada\\
  \texttt{mochen@cs.sfu.ca} \\
   \AND
  Inna Sharf \\
  Dept. of Mechanical Engineering \\
  McGill University \\
  Montreal, QC, Canada\\
  \texttt{inna.sharf@mcgill.ca} \\
}

\begin{document}
\maketitle

\begin{abstract}
 In this work, we propose a novel shared autonomy framework to operate articulated robots. We provide strategies to design both the task-oriented hierarchical planning and policy shaping algorithms for efficient human-robot interactions in context-aware operation of articulated robots. Our framework for interplay between the human and the autonomy, as the participating agents in the system, is particularly influenced by the ideas from multi-agent systems, game theory, and theory of mind for a sliding level of autonomy. We formulate the sequential hierarchical human-in-the-loop decision making process by extending MDPs and Options framework to shared autonomy, and make use of deep RL techniques to train an uncertainty-aware shared autonomy policy. To fine-tune the formulation to a human, we use history of the system states, human actions, and their error with respect to a surrogate optimal model to encode human's internal state embeddings, beyond the designed values, by using conditional VAEs. We showcase the effectiveness of our formulation for different human skill levels and degrees of cooperativeness by using  a case study of a feller-buncher machine in the challenging tasks of timber harvesting. Our framework is successful in providing a  sliding level of autonomy from fully autonomous to fully manual, and is particularly successful in handling a noisy non-cooperative human agent in the loop. The proposed framework advances the state-of-the-art  in shared autonomy for operating articulated robots, but can also be applied to other domains where autonomous operation is the ultimate goal. 
\end{abstract}

\keywords{Shared Autonomy \and Human-Robot Interaction \and Hierarchical Planning \and Policy Shaping \and MDP \and Deep RL \and cVAE \and Articulated Robots.}

\section{INTRODUCTION} \label{sec:intro}
    \subsection{Background and Motivation} \label{sec:motive}
        Shared autonomy is a framework to enable humans and robots to interact in a shared manner in order to accomplish certain goals. Shared autonomy has been utilized in a wide range of applications, from autonomous driving (\cite{Kiran2021}) to assistive robots (\cite{Losey2022}), in order to extend and enhance human capabilities (\cite{IEEE_CSS_2023RoadMap}). Indeed, the wide range of its applications attests to the importance of efficient human-robot interactions, as well as the current state of co-existence between humans and increasingly more intelligent robots.
        
        Our interest in shared autonomy is motivated by its potential applications in the context of articulated machines, these commonly operated by a human operator physically located in the machine. Operating such a machine, typically comprised of a mobile base and a large-scale multi-degree-of-freedom arm, involves multiple levels of hierarchy in the operator decision making. These encompass the higher-level strategical decision making, such as path planning for the machine,  all the way to the lower-level decision-making related to  individual joint control for arm manipulation. In essence, this hierarchy is comparable to human decision making when driving a car (\cite{Guo2019}). 
        \begin{figure}[!tbh]
            \centering
            \includegraphics[width=4cm]{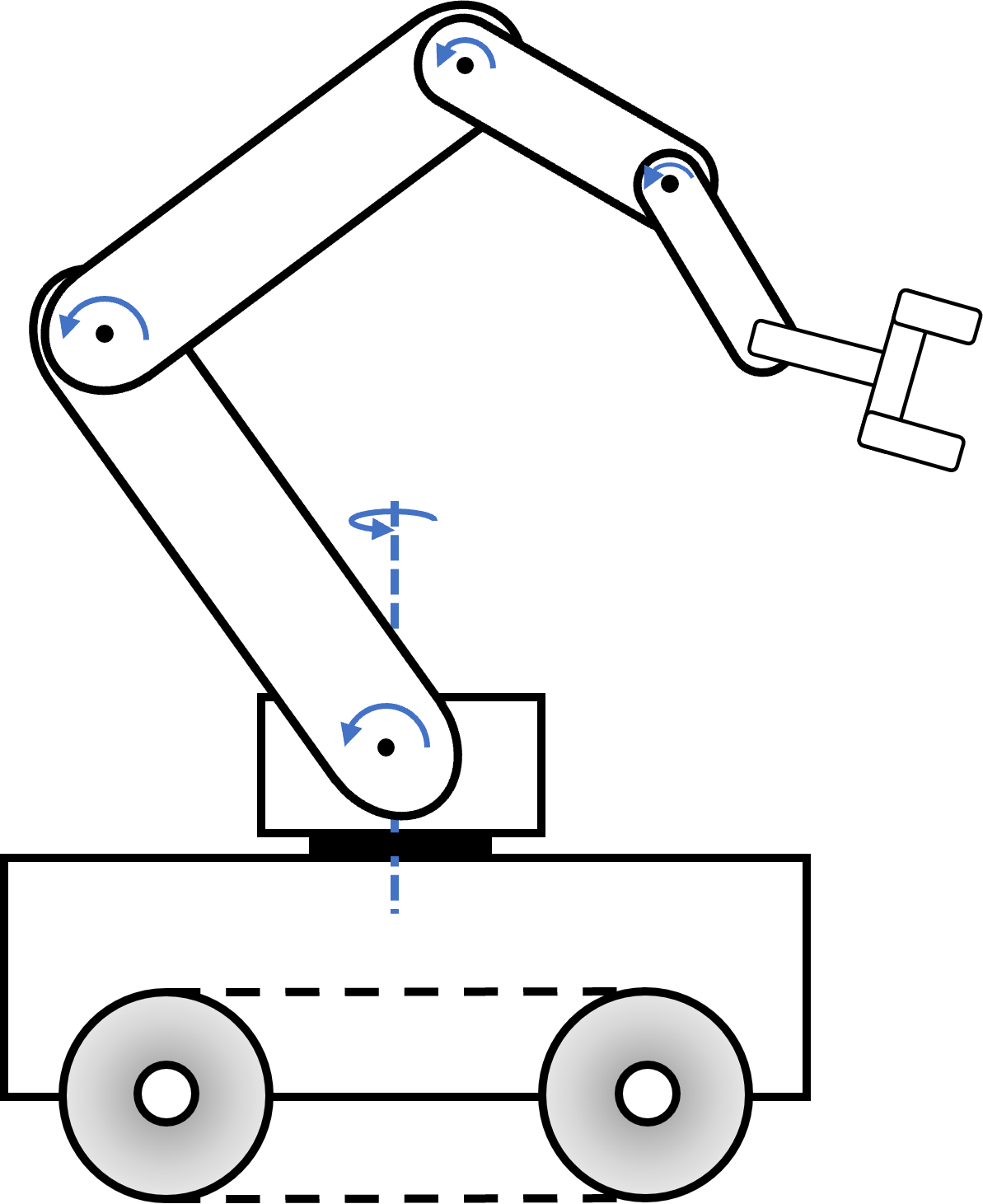}
            \caption{Schematic of an articulated robot with mobile base.}
            \label{fig:cute_robot}
        \end{figure}
         A schematic of an articulated robot with mobile base is shown in Figure \ref{fig:cute_robot}. 
         
         In addition to the hierarchy of decision levels as described above, operation of articulated machines, especially those in industrial settings,  such as excavators used in construction or feller-bunchers employed in timber harvesting, is also tied to the detailed know-how of their respective application domains. This is one of the reasons why reaching a high operator skill level to efficiently utilize these machines can take years in some applications (\cite{Westerberg2014, Lofgren2009}).
        In this paper, we develop a general task-oriented hierarchical planning framework for the robot/machine, 
        with human and AI interactions in mind, that extends beyond the standard robotic planning techniques.
        
        One of the main challenges  in the multi-level, real-world robotic applications is that despite having a good insight into different operations, complete knowledge of the relevant application domain needed to achieve a fully autonomous system cannot be assumed. However, this should not prevent us from incorporating whatever knowledge we have into a versatile framework. Moreover, the type of applications considered here, involving large, extremely powerful machines, does not allow for hazardous trial-and-error experimentation in the field. This motivates the central research question addressed in this work, that of how to design a comprehensive shared autonomy architecture that allows different levels of autonomy in a human-in-the-loop framework for complex, hierarchical, robotic decision-making tasks. It is important to highlight that in this work, the agents, i.e., the human and the autonomous, \textit{co-operate} on one physical entity, the robot/machine. Once again, this bears similarities to autonomous driving scenarios (\cite{Amini2020}), and is unlike many other human-robot interaction scenarios where the two agents act on/through two separate physical entities (\cite{hong2023learning,Dragan2017planning}).

    \subsection{State of the Art} \label{sec:stateofart}
        In shared autonomy, \textit{arbitration} of human and autonomous action commands, which jointly form the input to the robot/machine system, is of prominent importance. In this regard, the available schemes in literature  can be categorized into two main groups. The first is referred to as \textit{policy blending}, where the human action and autonomous action are treated as two separate signals and an \textit{arbitration function} is used to decide how to blend these two signals (\cite{Dragan2013}). Despite wide application  due to its  simplicity and efficacy, the policy blending approach has some drawbacks that stem  from the fact that it attempts to blend two signals that might be different in nature and their meaning (\cite{Javdani2018}). To address the latter issue, in (\cite{Losey2022}), the authors suggested a latent-action representation from human's low-dimensional actions to high-dimensional inputs. They next combined the latter with the assistance signal in order to fine-tune the robot behavior.
        The other limitation  is the inherent ``predict-then-go" nature of the system architecture to implement the policy blending approach (\cite{Javdani2018}). 
        In some respects, the resulting autonomous agent in an inherently ``predict-then-go" setting can be viewed as a Sisyphus or an \textit{absurd} hero (\cite{camus2018myth}), with a perpetual though successful struggle, that resets after every cue by the human agent.
     
        The second group of methods is what we will call \textit{policy shaping}, where the autonomous action (and policy) is shaped by taking into account human action, as well as other available information, and it is the \textit{only} input to the robot. In other words, unlike policy blending where the agents' inputs (i.e., human and  autonomous) are combined in parallel, in policy shaping, they are in series. This approach does not suffer from the drawbacks of policy blending;  however, it is computationally expensive and users report less comfort using it despite having better performance in certain scenarios (\cite{Javdani2018, ReddyDraganLevine2018}).
    
        One of the strategies in the second category involves conditioning the robot action on the human signal. The authors of (\cite{Javdani2018}) defined an augmented (autonomous) state consisting of the (overall) state of the robot and \textit{human's goal}. It was assumed that the human policy that acts based on the augmented state is modeled and known, for which they used the Maximum Entropy (MaxEnt) Inverse Optimal Control (IOC) framework. The autonomous action is based on the overall robotic system state as well as the human action, and is defined such that it minimizes a cost function dependent on the human action and goal. It was furthermore assumed  that a goal $g$ is partially observable and the human state is the same as the autonomous state. In (\cite{ReddyDraganLevine2018}), the authors developed a deep Reinforcement Learning (RL) algorithm to learn a model-free policy that maps the augmented state of the robot to the (autonomous) action. The augmented state comprised the state of the overall robotic system and the \textit{human signal}. The latter was either the intended goal -- inferred using Bayesian inference under an inverse RL scheme -- if such information was available, or the raw low-level human inputs, otherwise. The purpose in (\cite{ReddyDraganLevine2018}) was to find an optimal autonomous action close to the human action to deliver high performance, while keeping the human as a high-quality input source in the loop. It was demonstrated  that incorporating an inference algorithm resulted in a better overall performance despite the additional computational cost. However, the authors of (\cite{ReddyDraganLevine2018}) did not assume a model for human policy and the human signal was  part of an augmented state definition for the autonomous policy. A model-free RL algorithm was used to find the optimal autonomous action while keeping it close to the human action. 
        
        \subsection{Contributions} \label{sec:contributions}
            Our work is based on the premise that the ultimate goal of a fully autonomous system operating an articulated robot/machine is best achieved through a shared autonomy framework. Under such a framework, the \textit{autonomous agent} can progressively increase the level of autonomy while keeping the human in the loop to handle edge cases and to, possibly, learn from or teach the autonomous agent. We suggest that such a framework is  particularly useful to applications which rely heavily on a skilled human to operate the robot/machine, when the operations involve a hierarchy of decision making, and in operations where safety is important. With this perspective, the main contributions of this paper are as follows: 
            \begin{enumerate}[i)]
                \item Development of a general, task-oriented hierarchical planning formulation for the operation of articulated robots/machines, with human interpretability and shared autonomy in mind;
                \item Proposition of a novel shared autonomy architecture for human-in-the-loop tasks and policy shaping; this involves a design of hierarchical interactions and \textit{arbitration} between the autonomy and the human. Our work towards this contribution is particularly influenced by the ideas from multi-agent systems, game theory, and theory of mind (\cite{pynadath2005psychsim,Dragan2017planning}) for a sliding level of autonomy;
                \item Formulation of the MDPs and Options framework to enable deep RL for shared autonomy;
                \item Application of the  proposed shared autonomy framework to an industrially important application---timber harvesting. Thus, we fine-tune our formulation for the specific tasks of a timber-harvesting machine:  a feller-buncher, which is a large-scale hydraulically actuated articulated robot with a specialized end-effector (\cite{Yousefi2022}). 
            \end{enumerate}
        
            This paper is organized as follows: We first introduce the definitions and nomenclature in \S\ref{sec:defs}. In \S\ref{sec:elements}, we provide the elements of shared autonomy. Then, in \S \ref{sec:problem_statement}, we provide our problem statement and points of view on the problem. In \S\ref{sec:application}, we discuss the case study application: timber harvesting, followed by detailed analysis in \S\ref{sec:shared_FB}. In \S\ref{sec:results}, we present our results for different sections. Finally, \S\ref{sec:conclusion} concludes our work by reiterating the main ideas presented and suggesting directions for future work.

\section{DEFINITIONS AND NOTATION} \label{sec:defs}
    \begin{figure}[!tbh]
        \centering
        \includegraphics[width=10cm]{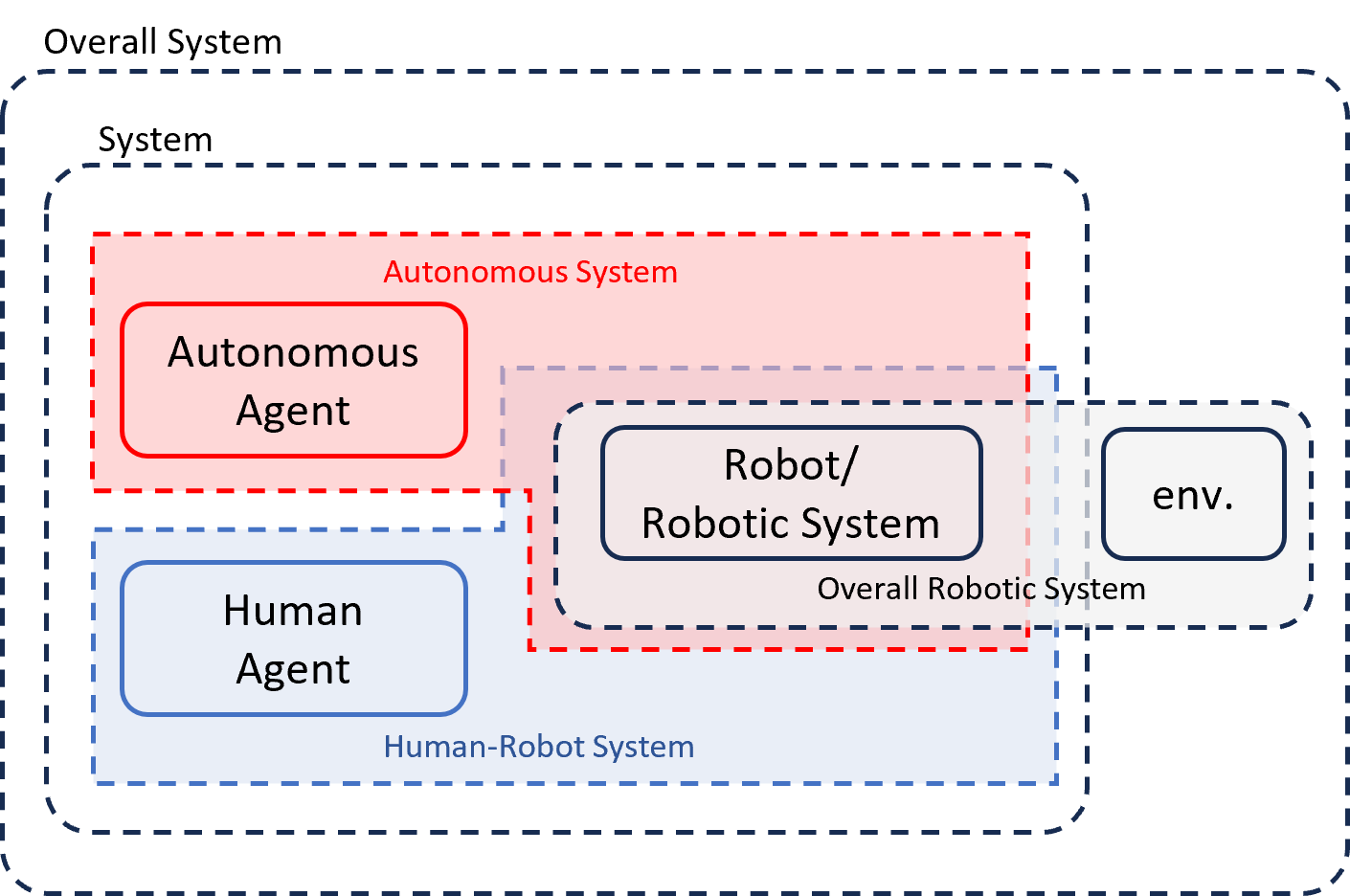}
        \caption{Definition of different terms in our work.}
        \label{fig:defs}
    \end{figure}
    To eliminate possible ambiguities and for most clarity, we begin by defining the relevant terminology in our work. As much as possible, we use terminology consistent with what is established in the relevant literature; however, we bear in mind that definitions often depend on the specific perspective and background of the authors.  Figure \ref{fig:defs} depicts graphically the various components of the system and the corresponding  terms to describe its components. 
    
        \noindent \textbf{Agent}: This term refers to any entity capable of making decisions. In our problem, we have two agents:
        \begin{itemize}
            \item {\it Human Agent (HA)}: This term refers to a human operator, driver, or user, as a decision maker.
            \item {\it Autonomous Agent (AA)}: This term refers to the artificial high-level intelligence capable of decision making.
        \end{itemize}
        \vspace{1mm}
        \noindent \textbf{Robot} or machine: the physical entity being operated by an agent; it is operated in the field and interacts with the \textit{environment}. In our application, it is a mobile base with an articulated arm, such as a feller-buncher machine operated in the forest. We might use the term \textit{robotic system} equivalently, as a robot includes certain components internally, such as sensors and actuators. We use the term \textit{Overall Robotic System} when we refer to the robotic system and the environment together.
        
        \vspace{2mm}
        
        \noindent \textbf{Autonomous System}: This term refers to the autonomous agent (AA) and the robot together. The term \textit{Overall Autonomous System} is used when we include the environment as an element in this system.
        
        \vspace{2mm}
        
        \noindent \textbf{Human-Robot System}: This term refers to the human agent (HA) and the robot together. The term \textit{Overall Human-Robot System} is used when we include the environment as an element in this system.
        
        \vspace{2mm}
        
        \noindent \textbf{System}: This term refers to the human agent (HA), the autonomous agent (AA), and the robot together. The term \textit{Overall System} is used when we include the environment as an element in this system. 

        With the above system components clearly delineated, we next introduce the basic terminology for the learning aspect of the framework.
        
        \vspace{2mm}        
        
        \noindent \textbf{State}: relevant variables defined for each element of the Overall System, in particular,

        \begin{itemize}
            \item State of the robot, $s^R$: This refers to variables relevant to the robot itself, such as its pose and the remaining capacity of its end effector, i.e., the end effector capacity for maneuverability ($CfM_{ee}$). 
            \item State of the environment, $s^E$: This state defines the different elements in the environment surrounding the robot, such as the objects, and obstacles, as well as certain task-related elements, depending on the type of task. We will discuss these in more detail in the subsequent sections. 
            \item State of the Autonomous Agent, $s^A$: This is the {\it designed} representation of the state of the Overall System by the architect of Autonomous Agent based on the foregoing state elements as well as task-related elements. This representation forms the basis upon which the autonomous agent acts.
            \item State of the Human Agent, $s^H$. This refers to the human's representation of the state of the Overall System. We do not assume knowledge of $s^H$. Also, we do not assume equivalency between $s^H$ and $s^A$ as will be discussed later.
            \end{itemize}
        
        \vspace{2mm}
        
        \noindent \textbf{Action}: Each of the decision making agents, i.e., HA and AA,  can also \textit{act} in a shared autonomy setting depending on the collaboration scheme and level of autonomy. The action is relayed directly to the robot as input. We will use the following terms:
        \begin{itemize}
            \item Action of Autonomous Agent or simply \textbf{Autonomous Action}, $a^A$: In a shared autonomy setting, this action will be of assistive nature, and we might refer to it as \textit{Assistive Signal}.
            \item Action of Human Agent or simply \textbf{Human Action}, $a^H$: This refers to the human action using any input device, such as, joysticks and pedals.
        \end{itemize}
        
        \vspace{2mm}
        
        \noindent \textbf{Policy}: Each of the decision-making/acting agents in a shared autonomy setting has a policy according to which they act. We will use Human (Agent) Policy, $\pi_H$, and Autonomous (Agent) Policy, $\pi_A$, to refer to these policies.

\section{Elements of Shared Autonomy} \label{sec:elements}
    \subsection{Hierarchical task-oriented robot planning \& design variables} \label{sec:pp_gen}
    
        As alluded earlier, we consider the robot planning problem in terms of tasks and functions:
        this is advantageous when interfacing multiple agents, including a human in the context of shared autonomy, as well as sliding levels of autonomy (\cite{Khatib2022}). Moreover, this task-oriented approach makes it possible to incorporate the inherent hierarchy of tasks and consequently, hierarchy in human-robot interactions (\cite{Guo2019}). 
        \begin{figure}[!tbh]
            \centering
            \includegraphics[width=8cm]{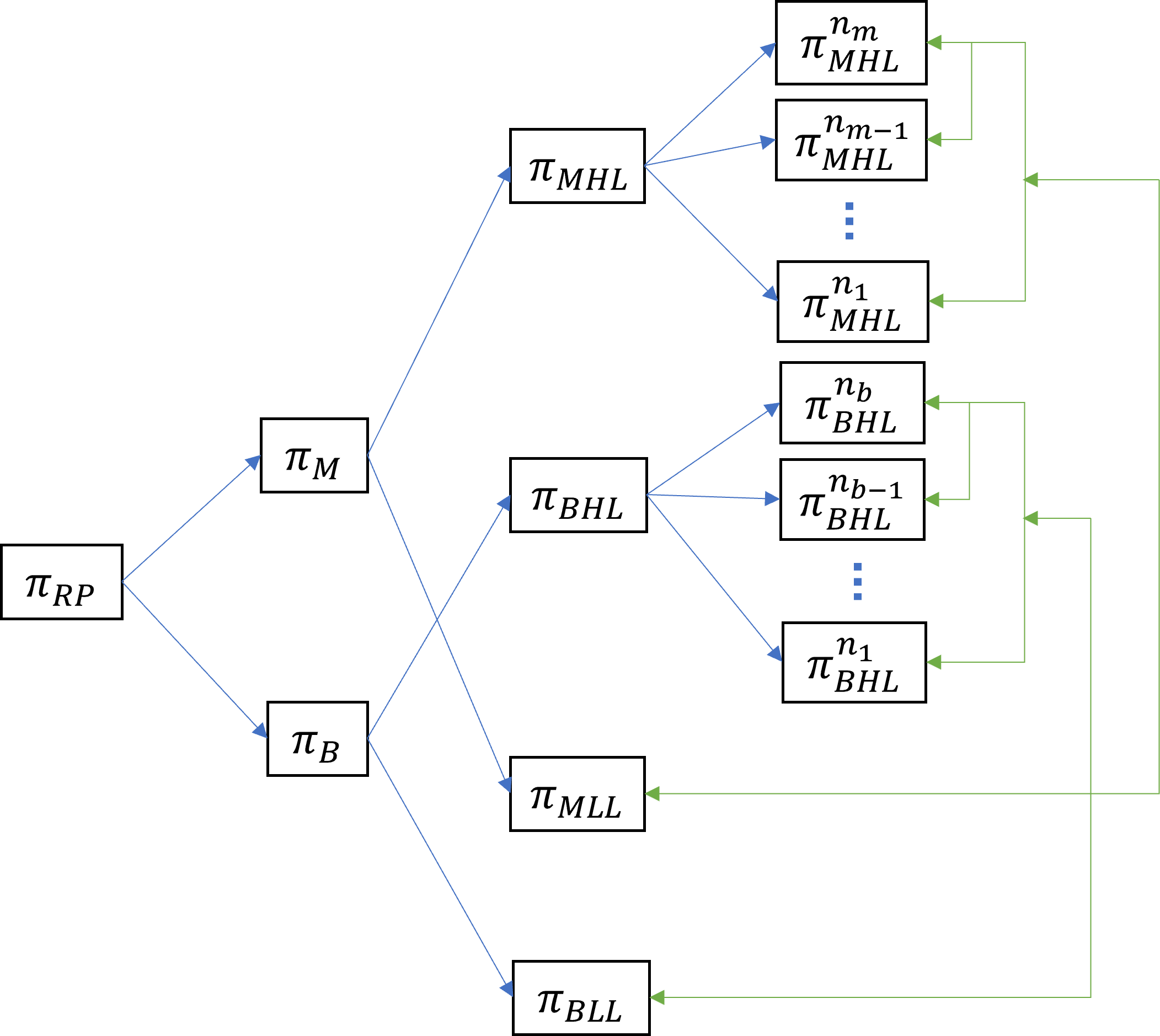}
            \caption{Hierarchical task-oriented breakdown (black arrows) of robot planning for articulated robots. The green arrows in Figure \ref{fig:RP_gen_breakdown} show the inter-dependencies of the (sub-)policies.}
            \label{fig:RP_gen_breakdown}
        \end{figure}
        \begin{figure}[!tbh]
            \centering
            \includegraphics[width=12cm]{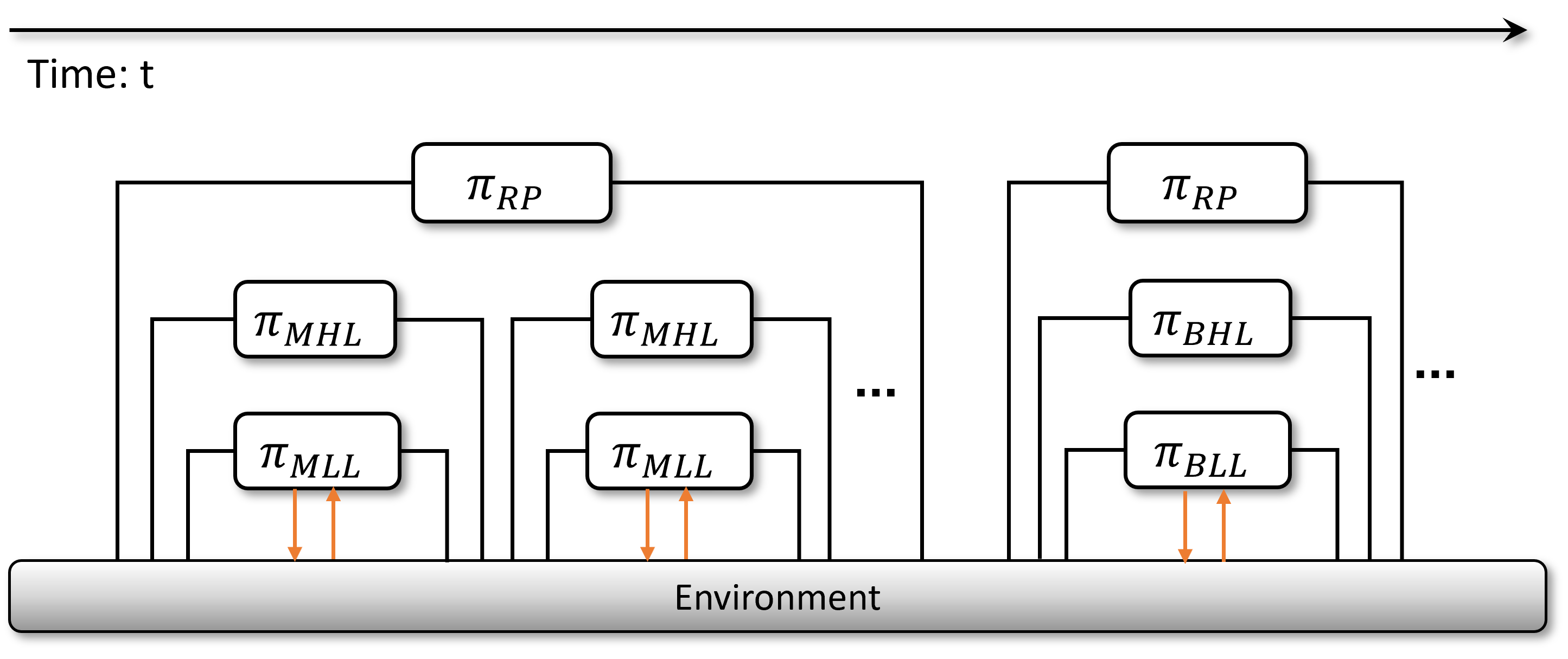}
            \caption{Temporal abstraction of hierarchical robot planning policies. The policies repeat over time and with different time scales as shown with ``...".}
            \label{fig:RP_gen_spatiotemporal}
        \end{figure}
        \begin{figure}[!tbh]
            \centering
            \includegraphics[width=6cm]{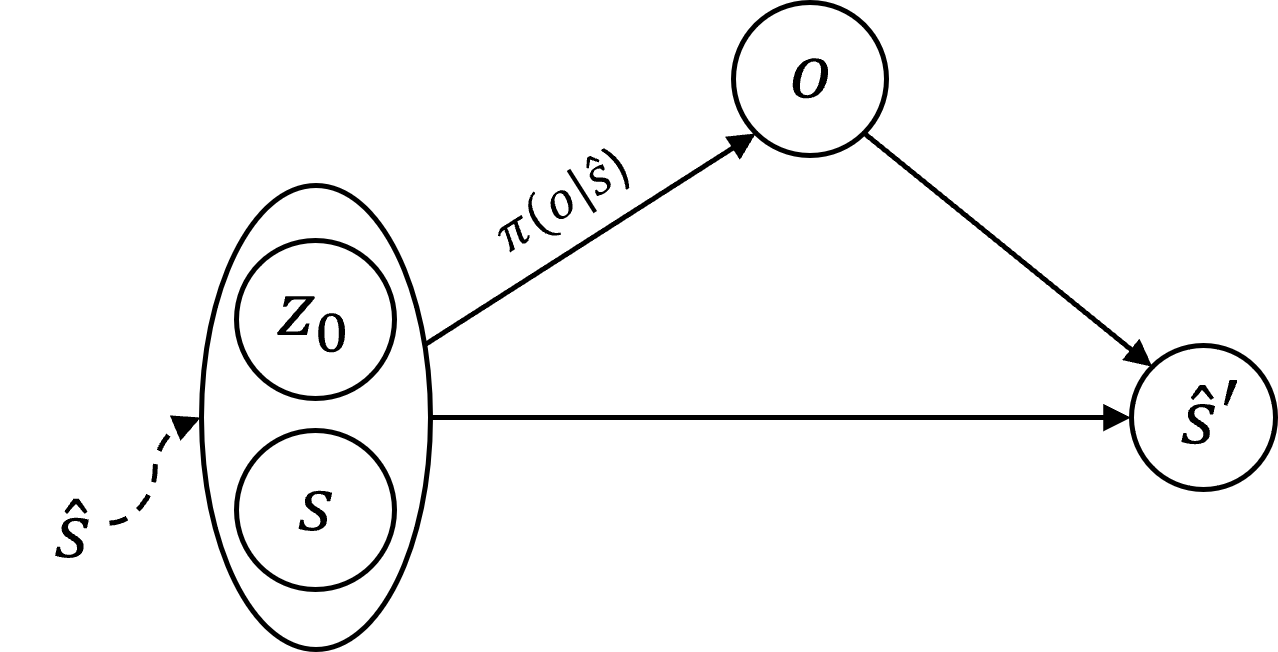}
            \caption{Graphical probabilistic model for a policy in task-oriented robot planning}
            \label{fig:sharedRL_layer_gen}
        \end{figure}
        
        The conceptual representation of our task-oriented hierarchical perspective on the robot planning problem for an articulated robot/machine is shown in Figure \ref{fig:RP_gen_breakdown}. Here, $\pi_{RP}$ denotes the overarching, master policy for robot planning and it is  broken down (black arrows) into two general tasks or policies and their associated  sub-policies as follows:
        
        \noindent {\bf $\pi_M$:} {\it policy to move the arm.} This is further categorized into two hierarchical levels:
        \begin{itemize}
            \item $\pi_{MHL}$: policy for high-level arm manipulations that includes $n_m$ sub-policies, such as end-effector path planning and scheduling of arm motions. The specific definition of each sub-policy depends on the particular application domain beyond standard robotic planners;
            \item $\pi_{MLL}$: policy for low-level arm manipulations that includes low-level control, e.g., joint control;
        \end{itemize}
        
        \noindent $\pi_B$: {\it policy to move the base of the robot/machine.} This is further categorized into two hierarchical levels, defined similar to the arm motion policies:
        \begin{itemize}
            \item $\pi_{BHL}$: policy for high-level motion of the base that includes $n_b$ sub-policies, such as the classic example of hierarchical room-to-room robot planning (\cite{precup1997multi}),
            \item $\pi_{BLL}$: policy for low-level motion of the base that includes base motion control.
        \end{itemize}
        The green arrows in Figure \ref{fig:RP_gen_breakdown} show the inter-dependencies of the (sub-)policies. It should be noted that there might be policies that require coordinated or combined  planning of robot arm and base. This would also fall under the umbrella of the overarching robot planning policy.
        
        A task-oriented planning (and scheduling) problem can be formulated as a sequential decision making problem that optimizes for certain task-specific  metrics (\cite{Yousefi2022}). We use the analogy between an \textit{option} in Options framework (\cite{Sutton1999, Pateria2021}) and a task in our robot planning: just like a task may involve multiple sub-tasks, an option generally involves multiple actions. The Options framework encodes the generalized actions as options. With this analogy, we invoke the Markov decision process (MDP) framework which  provides a model for sequential decision making processes, considering the agents' actions while taking into account the stochasticity of the process. Since the Options framework itself is built on semi-MDPs which extend the definition of MDPs to include a sense of \textit{time}, it enables our shared autonomy formulation to accommodate robotic tasks with different durations, as well as hierarchies. 
        Moreover, it has been shown that the behavior of a human agent operating an articulated robot can be described by a \textit{well-structured sequence} of repetitive (sub-)tasks (\cite{Westerberg2014}). Our framework, therefore, is designed to take into account the \textit{spatiotemporal} aspects of a \textit{shared} policy in providing the abstraction of shared autonomy. An example of temporal progression of a sequence of hierarchical tasks is shown in Figure \ref{fig:RP_gen_spatiotemporal}.
        
        From a graphical probabilistic model point of view, the policy of an autonomous agent can be depicted  as in Figure \ref{fig:sharedRL_layer_gen}, where $z_0$ encodes the \textit{task} or \textit{function} specific state variables, which augment  the classic robot-related states  $s$ into $\hat{s}$. Research on how to  define $z_0$ is quite extensive and is also application specific, as illustrated by the work in (\cite{Losey2022, ReddyDraganLevine2018, dragan2013policy}). It could be argued that there is neither a unique formulation nor a methodology to define $z_0$, as there is no unique way of performing the same task. In tasks involving pick and place operations,
        using information about the goal space to define $z_0$ has been shown a good choice (\cite{Losey2022}). We will demonstrate the significance of this choice through an example in \S\ref{sec:results_for_robot_planning}. 
        The advantage of our framework is that $z_0$ is defined for shared autonomy by design, which makes the robot operation interpretable as well as efficient. The mutual interpretability attribute  is particularly important for tasks involving a human agent in the loop and for those with limited domain knowledge. It should be noted that our framework is not necessarily a ``human-knowledge-based" method\footnote{The Bitter Lesson, Rich Sutton, 2019: \\ \url{http://www.incompleteideas.net/IncIdeas/BitterLesson.html} (accessed 12.06.2023).}. Although having insight into how humans perform a task helps with the understanding of the task, especially for applications of  human-operated machines, this is not a requirement for our framework, but a matter of interpretability to the human who is in the loop.
        
        With the understanding presented above, we now discuss how we interface the agents in the system, i.e., the human and the autonomous, given the hierarchy of tasks. Our view of the overall system involving a human and an autonomous agent is that of  a multi-agent system with gamified interactions, and we design our shared autonomy architecture accordingly, as discussed next.        

    \subsection{Shared autonomy architecture \& design variables}
    
\begin{figure}[p!tbh]
            \centering
            \includegraphics[width=15cm]{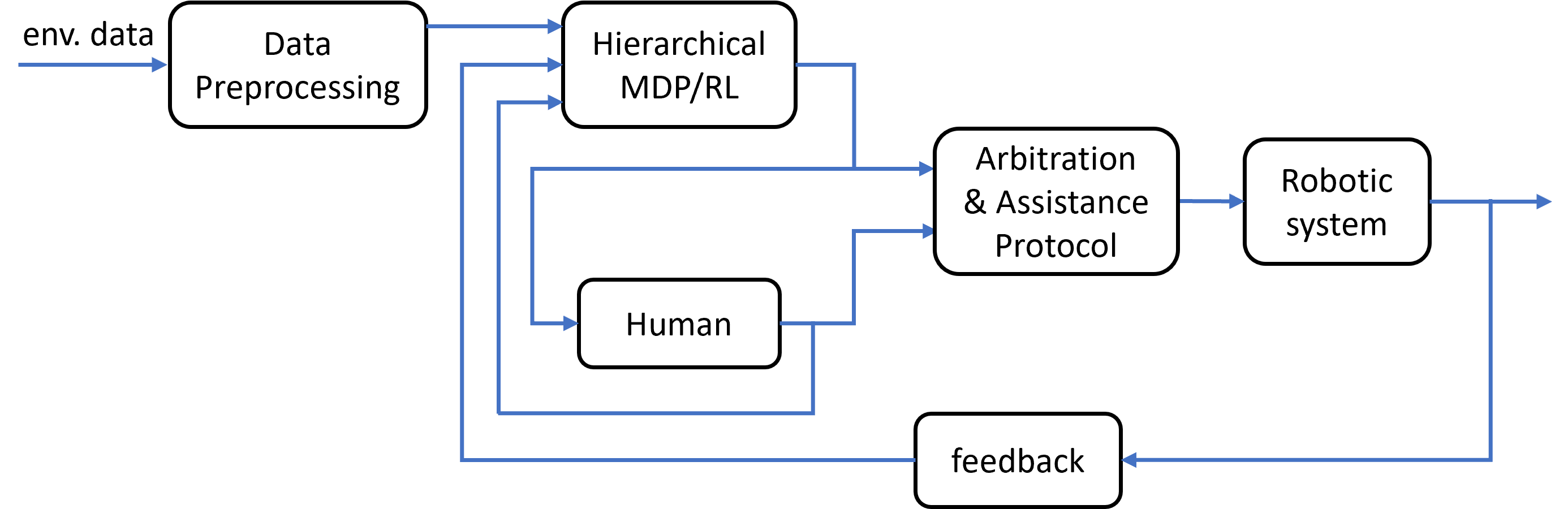}
            \caption{Block diagram of the general shared autonomy scheme}
            \label{fig:block_diag_shared}
        \end{figure}

\begin{figure}[!tbh]
            \centering
            \includegraphics[width=8cm]{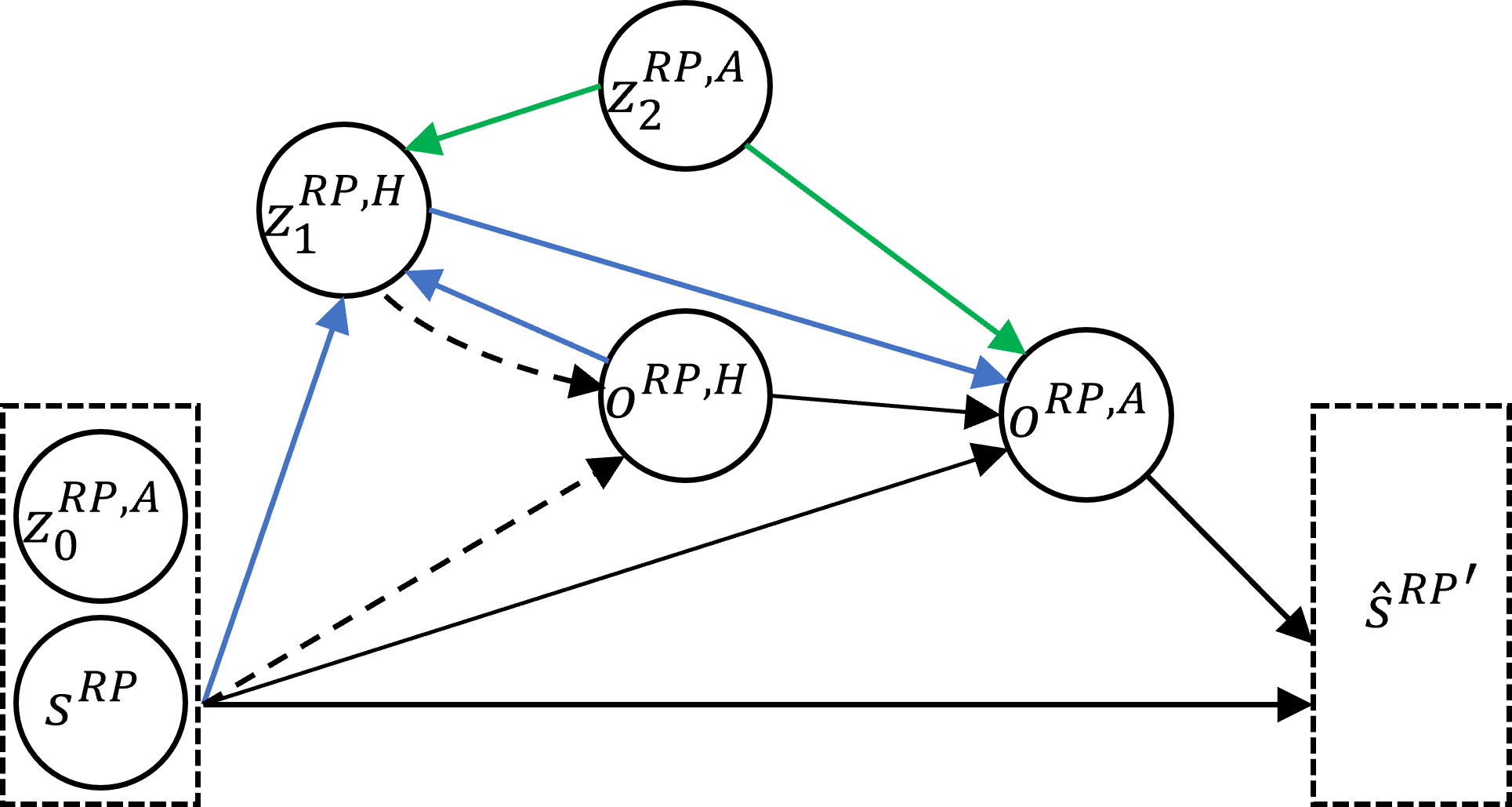}
            \caption{Graphical probabilistic model for shared autonomy. The dashed arrows are related to cVAE and human analysis discussed in \S\ref{sec:human_analysis}. The green arrows are related to including pre-trained model in the shared autonomy operation. The blue arrows designate the flow of signals to human and our inferred internal state variable, $z_1$. 
            }
            \label{fig:sharedRL_simp}
        \end{figure}

        A high-level block diagram of the proposed shared autonomy scheme from the control systems point of view is shown in Figure \ref{fig:block_diag_shared}. The architecture of each of these blocks and how they interface with each other are  the most challenging aspects of shared autonomy design. Figure \ref{fig:sharedRL_simp} depicts our proposed viewpoint of a shared autonomy framework as a graphical probabilistic model. With the task-representative variables, $z_0$, as introduced in \S\ref{sec:pp_gen}, we now introduce the human-representative design variables, $z_1$, which encode the human's internal states. 
        
        There is substantial literature on conceptualizing the human aspect in the context of shared autonomy, whether through explicit model assumptions (for example, (\cite{Dragan2013}) or model-free approaches (for example, (\cite{ReddyDraganLevine2018})). In the literature to date, inferred human's belief over the goals of the specific task is often used as one of the human internal states, even though inference of specific goals may not always be feasible. Arguably, optimality of the operator, or equivalently, the amount of noise in their actions, is another important characteristic that we are interested in quantifying and utilizing for smooth, user-tuned shared autonomy. Thus, human operator analysis with minimal assumptions about them is an important aspect of our work, to be discussed in \S\ref{sec:human_analysis}. The flow of information to/from a human agent is shown with blue arrows in Figure \ref{fig:sharedRL_simp}. The dashed lines are related the human analysis process that will be discussed in \S\ref{sec:human_analysis}.
        
        Continuing with Figure \ref{fig:sharedRL_simp}, we introduce $z^{RP,A}_2$ to encode \textit{pre-training} related state variables to represent prior  training and knowledge. In our framework, we refer to pre-training as the process of training a fully autonomous agent using a model similar to Figure \ref{fig:sharedRL_layer_gen}. This is the first step in shared autonomy design, ensuring  that the state definition and the model are capable of performing a task autonomously. As well, the pre-trained model will allow us  to look into the human's internal state and/or their fine-tuning and noise through the lens of a structured task. The information gained with this process enters the model via the  green arrows in Figure \ref{fig:sharedRL_simp}.
    
        To summarize, we employ three categories of variables in our shared autonomy framework:
         \begin{enumerate}
             \item $z_0$: task/operation specific state variables, representing the domain knowledge,
             \item $z_1$: human's internal states,
             \item $z_2$: pre-training  state variables. 
          \end{enumerate}
         These three categories can be considered as pillars of how  humans learn to perform a task: we bring in our past knowledge and experiences (category 3), fine-tune those for a particular series of tasks towards optimality based on the task requirements (category 1), and personalize how we proceed with taking actions (category 2). 
         
\begin{figure}[!tbh]
            \centering
            \includegraphics[width=8cm]{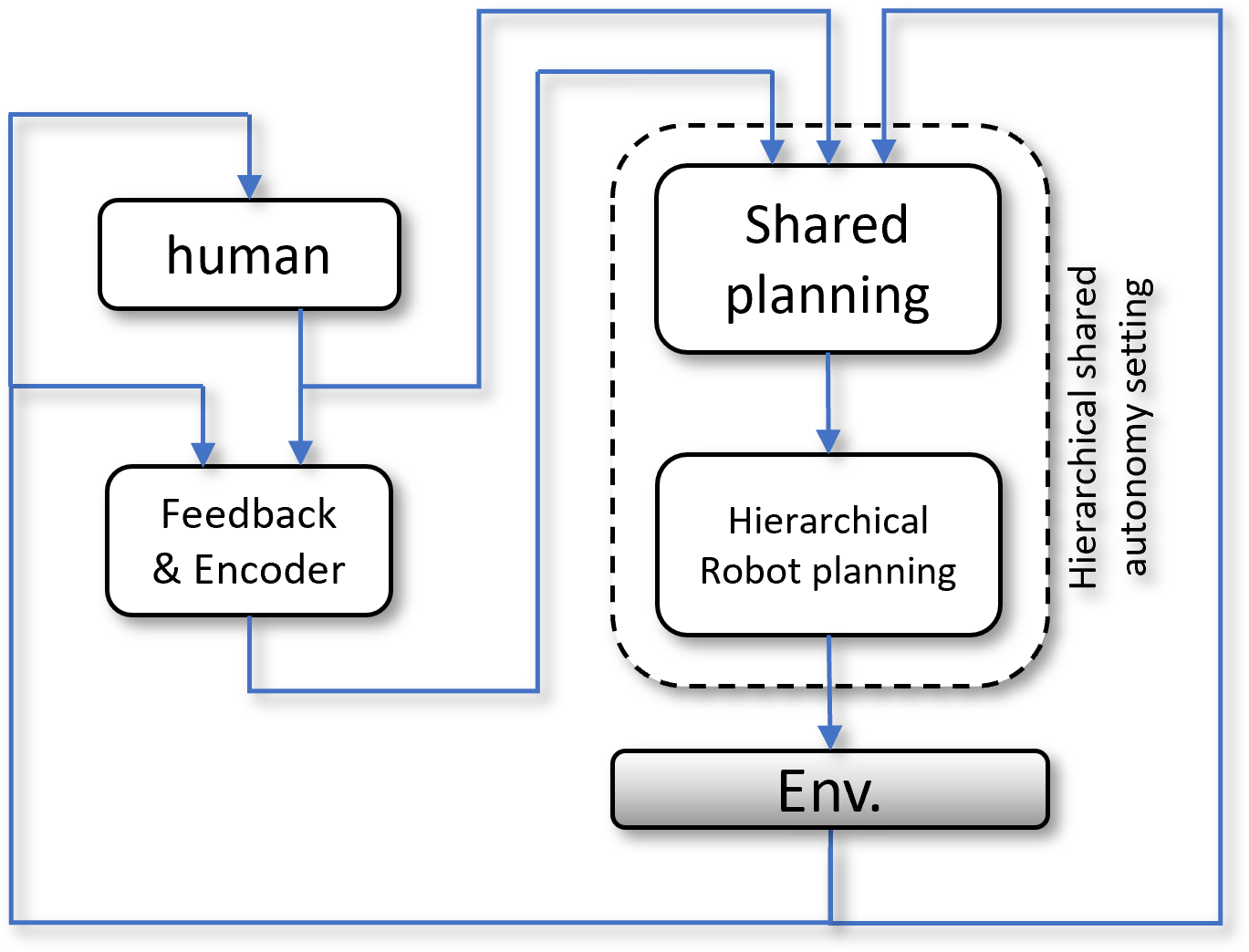}
            \caption{Simplified representation of hierarchical shared autonomy architecture}
            \label{fig:sharedRL_hierarchical_simp}
        \end{figure}

         In a shared autonomy context, this equivalence is helpful as it allows the system to be mutually understandable to both the human and the autonomous agents. Our proposed shared autonomy framework can also be considered as an instance of computational human-robot interaction (\cite{Thomaz2016}). The most important problem that we  address is how to design a hierarchical structure for shared autonomy so as to facilitate and further, to make seamless, this complex interaction of multiple hierarchical systems, as shown in Figure \ref{fig:sharedRL_hierarchical_simp}.
        
        In analyzing human behavior, it is important to note that, in general, to assume that the state definition $s_t$ is the same between the human agent and the autonomous agent is not valid. The reason lies in the fact that we do not have access to internal perception and state definition of a human, i.e., $s^H_t$. Therefore, assuming that a human policy maps from $s^A_t$ to $a^H_t$ is conceptually inaccurate, in general. In this paper, we assume that a human has an internal 
        state that is comprised of $s^A_t$ and $z_{1,t}$; hence, $s^H_t \triangleq (s^A_t, z_{1,t})$. The latter denotes the partially observable part of a human's state. This point becomes even more important when we have a hierarchy of tasks. The notion of including the human signal in the augmented state definition, as proposed in (\cite{ReddyDraganLevine2018}), makes sense in this light. If the human signal is defined as the low-level actions, then this implicitly enforces the Markovian assumption, i.e., the history is ignored. In contrast, it can be argued that conditioning the \textit{shared policy} $\pi_{sh}$ on a rich signal from the human, such as the goal space and if feasible the intended goal, is critical, as it effectively reflects the human history of actions. This argument is also supported by the results in (\cite{ReddyDraganLevine2018}) for an unstructured user input, where raw (i.e., unconditioned) low-level human inputs were used and poor performance was reported. Consequently, the performance of a collaboration scheme highly depends on a relatively successful encoding of human's internal state variable(s) $z_1$. It is worth noting that goal inference algorithms, in general, are based on a history of human inputs. The goal/intent inference requires a knowledge of goal space, which, in turn, requires domain knowledge. We encode the latter in $z_0$ without assuming direct knowledge of the intended goal, but only as a measure of goal space.
                
        From another point of view, using the human's internal signal as input to the autonomous policy provides a mechanism to \textit{synchronize} human actions and the resultant autonomous actions over a finite horizon that ends with reaching a goal. Otherwise, the performance of collaboration will be poor, as was the case in the results reported in (\cite{ReddyDraganLevine2018}) when low-level  human input was used in the autonomous policy shaping. 
\section{Problem Statement \& Modelling} \label{sec:problem_statement}
    We now present the mathematical model of our shared autonomy architecture in compact mathematical form. A typical trajectory $\tau$ of sequential state-actions in the context of human-robot interaction takes the following form:
    \begin{equation}
        \tau = \{s^A_t,a^H_t,a^A_t,...,s^A_T,a^H_T,a^A_T \}, \label{eq:dist_traj}
    \end{equation}
    where $s^A_t$, $a^A_t$, and $a^H_t$ denote the defined state, the autonomous agent action, and the human agent action, respectively, at any time-step $t$; $T$ denotes the time horizon for the task at hand. The action can be extended to an \textit{option} wherever needed. In this work, we do not impose a Markovian constraint on the human action, and thus, include a history of states in the human policy. Letting $n_h$ represent the number of steps of human's state history, it can be shown that the probability distribution of the trajectory is given by:
    \begin{equation}\label{eq:dist_traj2}
        \begin{split}
            p(\tau) &= p(s^A_1) \prod_{t=1}^{T}\pi_{H}(a^H_t|\overline{s}^A_t) \pi_{A}(a^A_t|a^H_t,s^A_t)p(s_{t+1}|s^A_t,a^A_t),
        \end{split}
    \end{equation}
    where $\pi_{H}$ and $\pi_{A}$ are human and autonomous policies, respectively.  The state variable $\overline{s}^A_t = \{s^A_t, ..., s^A_{t-n_h}\}$ comprises $n_h$ steps of history of state trajectory. Note that $t \geq n_h$. The derivation of \eqref{eq:dist_traj2} is given in Appendix A. We take a look at each of the terms in \eqref{eq:dist_traj2} in more details.
    
    \subsection{Analysis of Human Agent \& Policy $\pi_H$} \label{sec:human_analysis}
        \begin{figure}[!tbh]
            \centering
            \includegraphics[width=12cm]{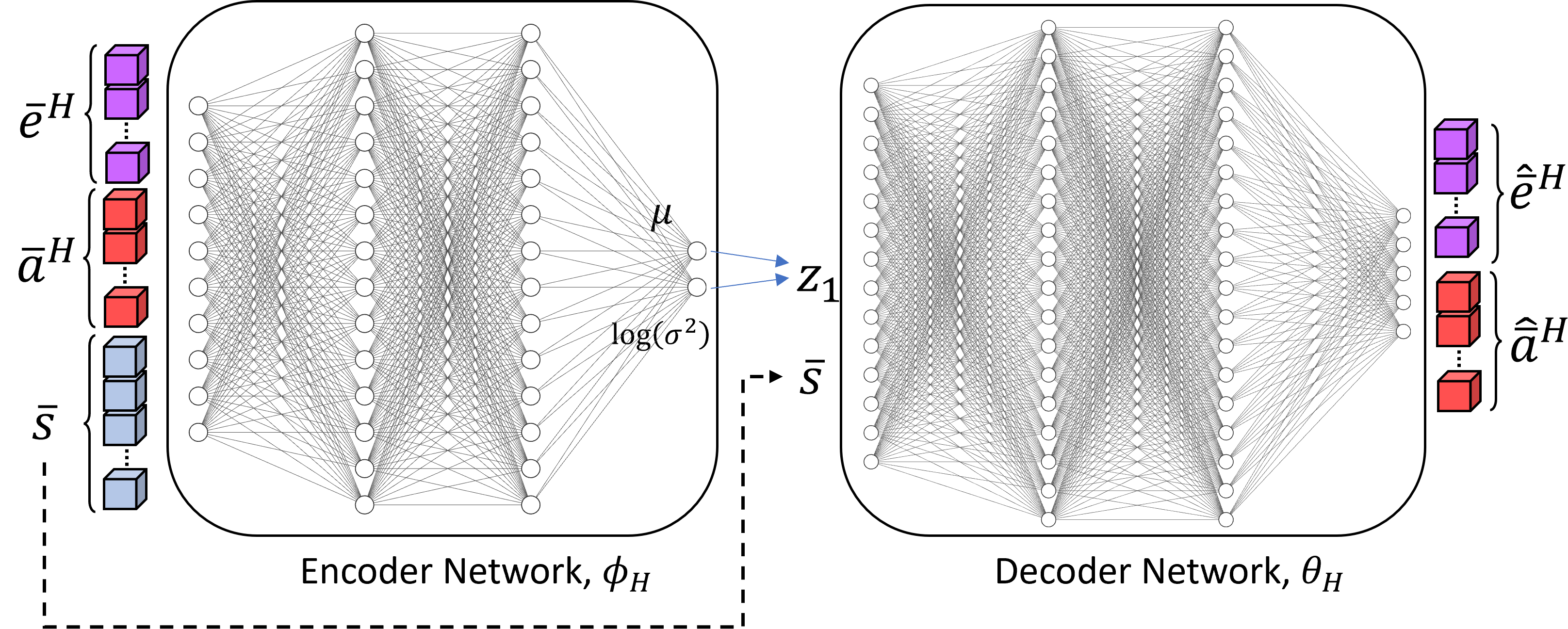}
            \caption{cVAE architecture to encode a measure of human performance in latent variable $z_1$}
            \label{fig:c_vae}
        \end{figure}
        As already noted, we do not assume equivalence between state definitions of the human and autonomous agents. Moreover, we do not assume any direct knowledge of the human's policy or their internal variables, $z_1$. 
        
        To address this knowledge gap and to analyze the $\pi_{H}(a^H_t|\overline{s}_t)$ term in \eqref{eq:dist_traj2}, we propose to explicitly encode the human's internal state variable $z_1$. This explicit encoding offers a deeper insight into the individual human agent; it helps to encode the differences between human agents and ultimately, enables a faster tuning of the shared autonomy framework to individual human operators. It also facilitates a more robust model against human \textit{noise} levels. We provide specific examples of this point in \S \ref{sec:results_for_sharedAutonomy}-\ref{sec:results_for_sharedAutonomy_part2}. 
        
        Let $n_s$ denote the dimension of autonomous state $\overline{S} \subset \mathbb{R}^{n_h \times n_s}$. We learn an encoder $\theta_H: \{\overline{E}, \overline{A}, \overline{S}\} \rightarrow \mathbi{Z}_1$, with human's latent state space $\mathbi{Z}_1 \subset d_{z_1}$, $d_{z_1} < (n_h \times n_s)$, from $\overline{S}$ conditioned on human's $n_h$ steps of history of actions $\overline{a}^H \subset \overline{A}$ as well as history of errors of their actions $\overline{e}^H \subset \overline{E}$ with respect to those of a known surrogate optimal agent,
        which might be another human or a pre-trained model. Considering $n_a$ discrete actions, $\overline{E} \subset \mathbb{N}^0$ and $\overline{A} \subset \mathbb{N}^0$. The error, in general, is defined as the angular difference between the denoted actions, as follows:
        \begin{equation}\label{eq:eH}
            \angle \mathbi{e}^H = \arccos(\mathbi{a}^H \cdot \mathbi{a}^*/\|\mathbi{a}^H\|\|\mathbi{a}^*\|),
        \end{equation}
        where $\mathbi{a}^*$ and $\mathbi{a}^H$ are the vectors representing the action of the surrogate optimal agent and that of the human, respectively. Moreover, we learn a decoder $\phi_H: \mathbi{Z}_1 \times \overline{S} \rightarrow \{\overline{A}, \overline{E}\}$, with the following reward function of the optimization process in cVAE defined as:
        \begin{multline} \label{eq:L_h}
            \mathcal{L}_{i,H} = E_{z_1 \sim q_{\phi_H}(z_1|\overline{e}^H_i,\overline{a}^H_i,\overline{s}_i)}(log p_{\theta_H}(\overline{e}^H_i,\overline{a}^H_i|\overline{s}_i,z_1)) - D_{KL}(q_{\phi_H}(z_1|\overline{e}^H_i,\overline{a}^H_i,\overline{s}_i)\parallel p(z_1)),
        \end{multline}
        where $\phi_H$ and $\theta_H$ are encoder and decoder networks, as shown in Figure \ref{fig:c_vae}. The two terms in \eqref{eq:L_h} are the reconstruction error and KL-divergence, respectively (\cite{Kingma_vae}).
        
    \subsection{Analysis of Autonomous Agent and Policy $\pi_A$} \label{sec:autnomous_agent}
        \begin{figure*}[!tbh]
            \centering
            \includegraphics[width=16cm]{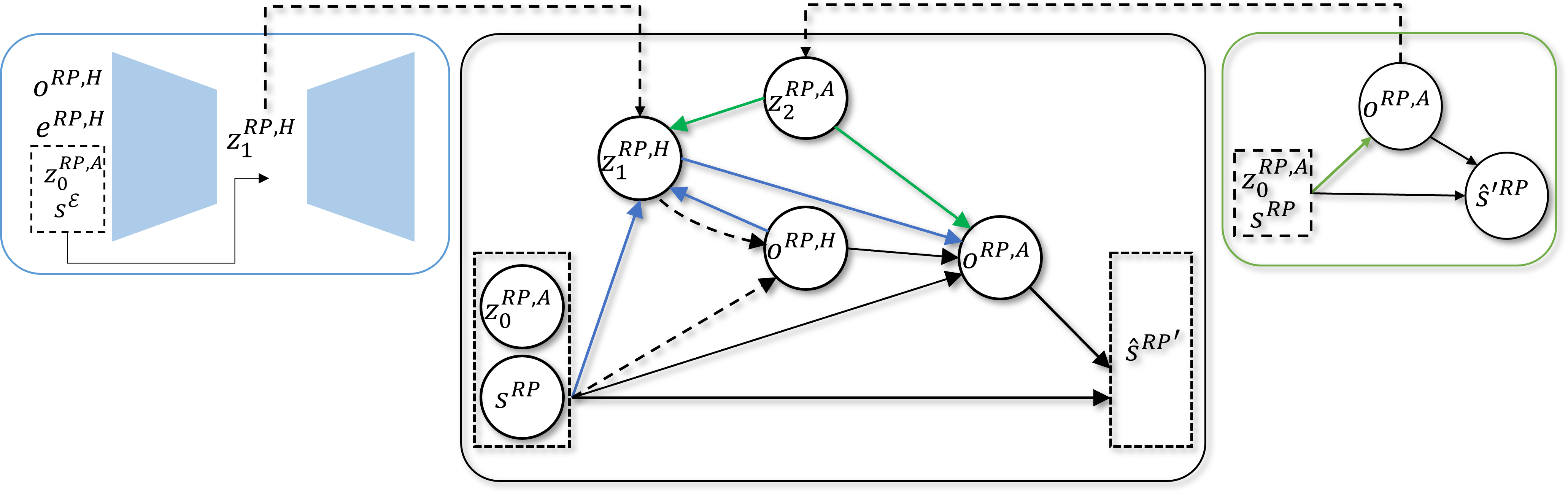}
            \caption{Hierarchical shared autonomy architecture for the complete graphical model of decision making. The middle box shows the main graphical probabilistic representation of the proposed shared autonomy framework. The box on the left shows the processing of human-tuned variable $z_1$, and the box on the right shows the processing of pre-training related variable, $z_2$.}
            \label{fig:h-shared-RL-arm}
         \end{figure*}
        Based on \eqref{eq:dist_traj2} and the discussions regarding encoding of distribution of $z_1$, the policy for the autonomous agent tuned to the  human agent can be written as:
        \begin{equation}\label{eq:autonomous_policy_formula}
            \pi_{A} = \pi_{A}(a^A_t|a^H_t,z_{1,t},s_t). 
        \end{equation}
        Following the logic of \S \ref{sec:human_analysis}, we now have access to the distribution of $z_1$. Based on \eqref{eq:autonomous_policy_formula}, we set up our shared autonomy framework, as shown in Figure \ref{fig:h-shared-RL-arm} in a graphical probabilistic model. It is worth noting that the degree to which a human agent participates in sharing the operation of the system depends on: (a) the level of autonomy desired for the system, (b) domain and task-specific knowledge, and (c) the extent of human presence. This shared autonomy framework facilitates a sliding level of autonomy. If we have a semi-autonomous agent, a shared autonomy framework is \textit{needed} to assist the human agent to reach a goal (\cite{Dragan2013}). The human agent's involvement, therefore, is in the training phase as well as testing/operational phases. 
        
        Moreover, we utilize the gamified human-robot interaction as well as game theoretic approaches in designing the reward function and the interaction architecture. The objective of this shared autonomy setting is to provide a near optimal input to the robot with respect to a reward function comprised of two contributions:
        \begin{itemize}
            \item $R_1$: Reward from robot planning, which includes task-related and obstacle avoidance rewards,
            \item $R_2$: Closeness to human input depending on signal $z_1$.
        \end{itemize}
    
        Hence, the general form of the reward function is as follows:
        \begin{equation} \label{eq:shared_reward}
            R = c_1R_1 + c_2R_2 
        = \mathbi{c}^T\mathbi{R}, 
        \end{equation}
        where $\mathbi{c}$ assembles the dynamic level of autonomy coefficients showing how much autonomy is required, how successful it has been, and in short, the level of autonomy. In other words, the choice of the two coefficients allows for a more efficient sliding level of autonomy. 
        
        From a multi-agent perspective, we model the agents' interactions and resolve possible issues in two ways: (1) policy shaping, that considers a serial architecture of the agents, and (2) strategic assignment of the coefficients $\mathbi{c}$. This is a novel perspective on the problem of multi-agent shared autonomy with human as an agent in the loop.
        \begin{figure}[p!tbh]
            \centering
            \includegraphics[width=8cm]{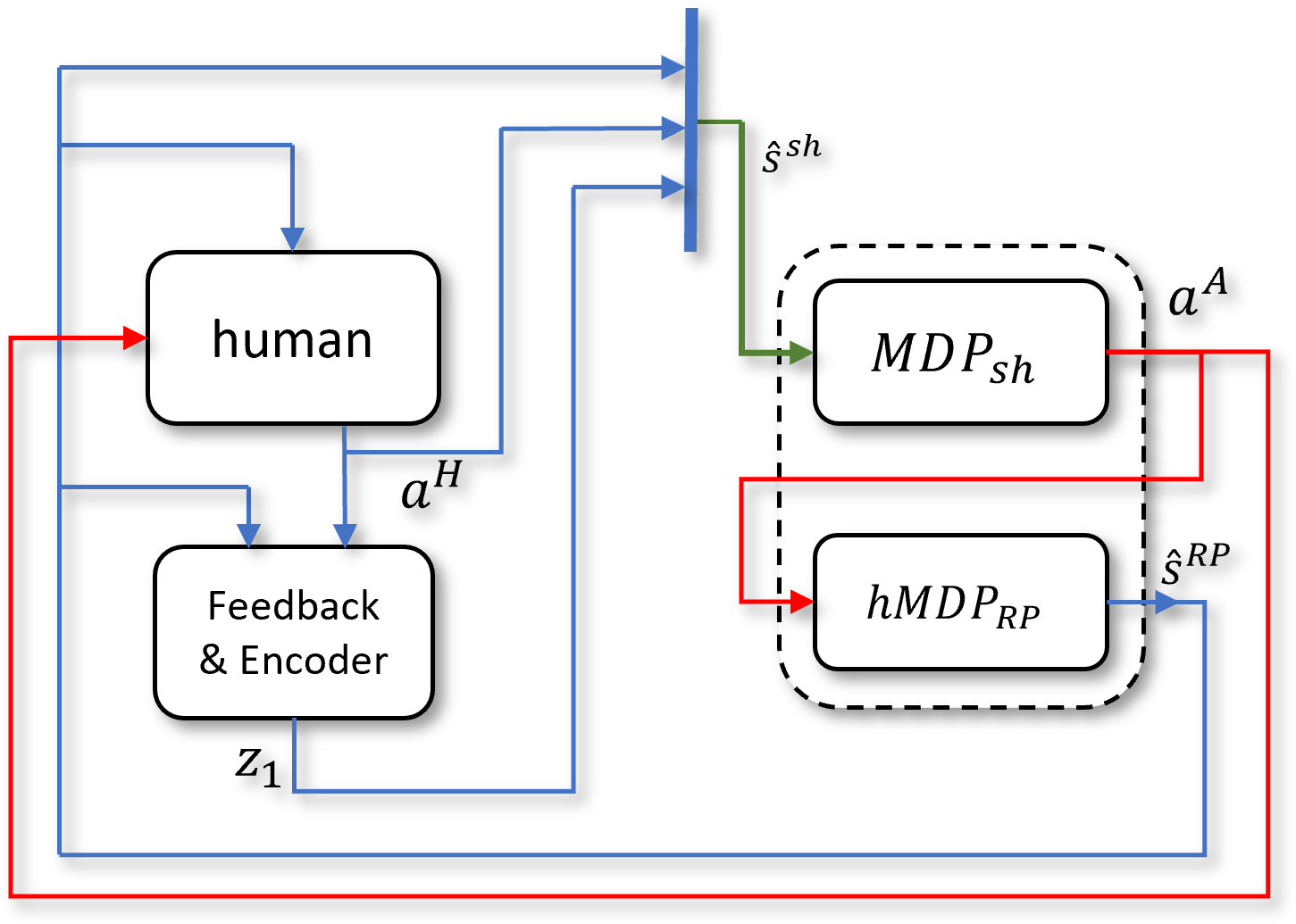}
            \caption{Complete representation of hierarchical shared autonomy architecture based on Figure \ref{fig:sharedRL_hierarchical_simp} utilizing hierarchical MDP (hMDP).}
            \label{fig:sharedRL_hierarchical_comp}
        \end{figure}            
        
       In summary, Figure \ref{fig:sharedRL_hierarchical_comp} shows our point of view on how the closed-loop block diagram of Figure \ref{fig:block_diag_shared} should be designed in a framework that is interface-able with the human as another agent for hierarchical robotic tasks.
       

\section{Case Study Application: Timber Harvesting} \label{sec:application}
    \begin{figure}[p!tbh] 
        \centering
        \includegraphics[width=6cm]{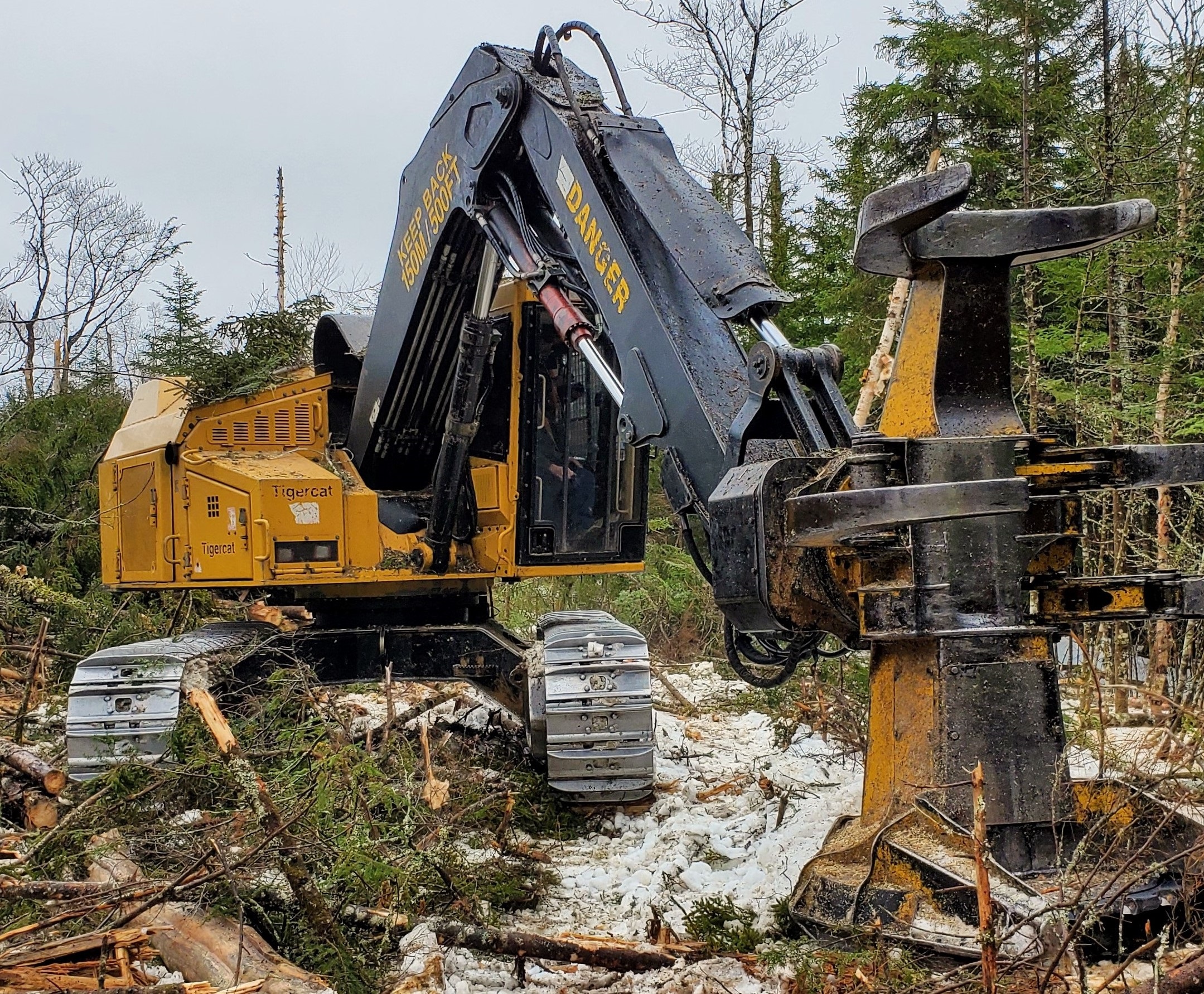}
        \caption{Example of robot/machine: a feller-buncher machine}
        \label{fig:fb_whole}
    \end{figure}
    \begin{figure}[tbh]
        \centering
        \begin{subfigure}[t]{0.404\textwidth}
            \centering
            \includegraphics[width=\linewidth]{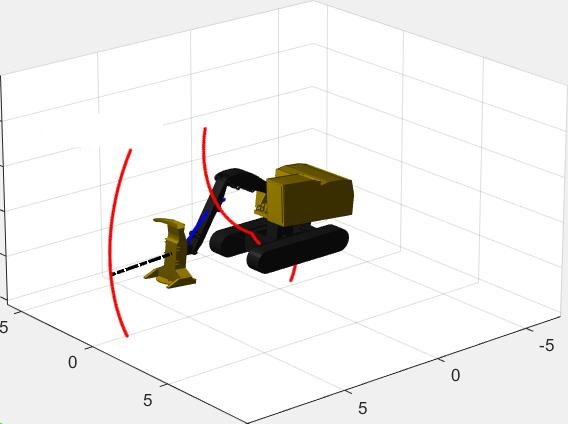}
            \caption{}
            \label{fig:fb_3d_P1}
        \end{subfigure}
        \hspace{0.2cm}
        \begin{subfigure}[t]{0.4\textwidth}
            \centering
            \includegraphics[width=\linewidth]{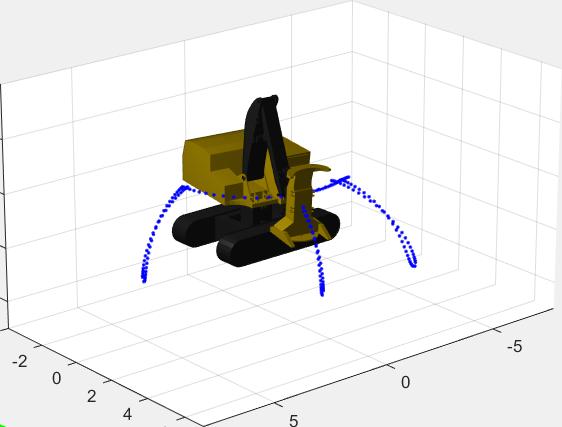}
            \caption{}
            \label{fig:fb_3d_P}
        \end{subfigure}
        \caption{(a) Robot performing option $O_1$ at a key point; (b) Robot end-effector trajectory on $E^M$ and to/from it during option $O_2$. $O_{1}$  encodes motion along the envelope from cell to cell, and $O_{2}$  encodes operations inside each cell.}
        \label{fig:fb_during}
    \end{figure}
    \begin{figure}[p!tbh] 
        \centering
        \includegraphics[width=8cm]{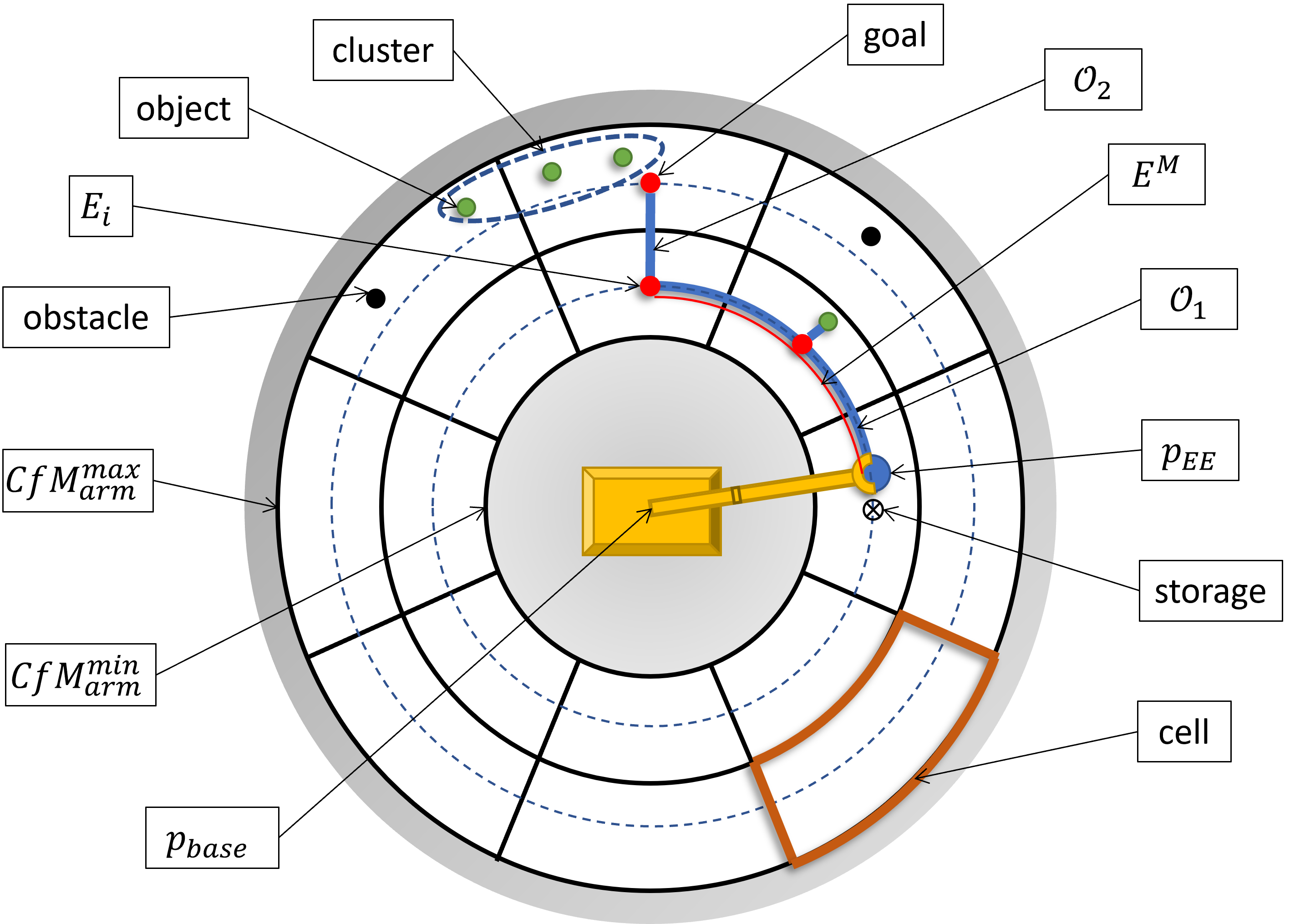}
        \caption{Setup for robot task planning with details.}
        \label{fig:pp_EM}
    \end{figure}
    \begin{figure}[p!tbh] 
        \centering
        \includegraphics[width=8cm]{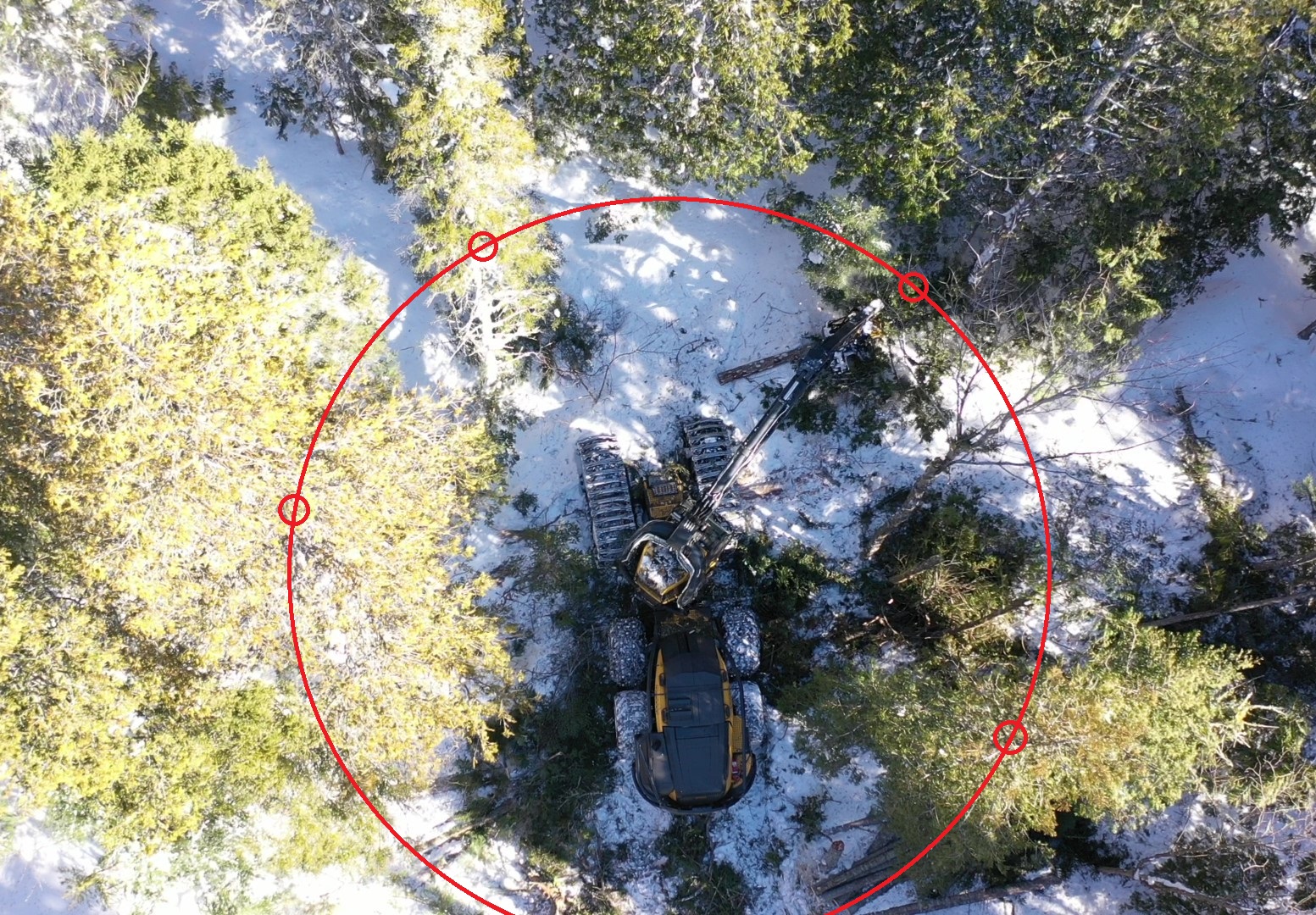}
        \caption{Top view photo of a harvester machine (similar in form to a feller-buncher) working in a forest site. The photo shows the analogy to our setup in Figure \ref{fig:pp_EM}.}
        \label{fig:top_view_site}
    \end{figure}

    As mentioned in \S \ref{sec:intro}, the motivation for our research on shared autonomy stems from its potential applications to machines employed in the Canadian timber harvesting industry. These machines, such as the feller-buncher machine used in our case study (see Figure \ref{fig:fb_whole}), are comprised of a mobile base and, a crane-like hydraulically-actuated manipulator arm with a specialized end-effector. In the case of feller-buncher, the latter is designed for cutting trees, picking them up and depositing them in a storage location.  Currently, machines employed for timber harvesting rely heavily on direct operator intervention, sometimes at the level of controlling individual joints of the crane. In fact, the current state of autonomy in the industry is much lower than in other comparable industries, such as mining (\cite{Lindroos2017,Fukui2017}). 

    There are several drivers for increasing autonomy of the machines employed in timber harvesting such as to improve the productivity of the operations which are significantly affected by human performance: for example, it has been reported that the productivity of a harvester is 25-40\% dependent on the skill of the operator (\cite{Lofgren2009}). In addition, the harsh machine and environmental conditions contribute to human fatigue, health issues and compromise operator safety, all of these factors exasperating the labour shortage for machine operators.  Moreover, it can take years of working in the field to achieve a skill level necessary  for operating the machine at a high level of productivity. This becomes apparent if one considers, for example, that the operator of a harvesting machine is required to perform, on average,  24 functions per tree and to make  12 decisions (\cite{Lofgren2009}). We suggest that shared autonomy can provide the way forward to address both the issue of productivity and operator training. The human-in-the-loop approach also addresses other challenges of complex robotic tasks, in particular the limited knowledge of their details and ensuring a certain level of safety.
    
    We consider the operation of a feller-buncher at a particular fixed location in the forest (i.e., fixed base), as  defined by the operation \textit{region} in the ground plane, illustrated in Figure \ref{fig:pp_EM}; the region is bounded by the minimum and maximum reach of the robot end-effector centered at the location of the mobile base. An actual photo of an operation region is shown in Figure \ref{fig:top_view_site} for comparison. We use the term Capacity for Maneuverability ($CfM_{arm} \in [0, 1]$), which is the remaining actuation capacity for a human intervention (\cite{eraslan2019shared}). A second capacity is defined for the end-effector since it can pick up several cut trees at a time: $CfM_{ee}  \in [0, 1]$ quantifies the remaining capacity in the end-effector to carry objects. With the view to describing the feller-buncher operation as an MDP, we divide each region into \textit{cell}s, which discretely encode the location of the machine end-effector $p_{EE}$, the objects in the region (i.e., trees), the goal location(s), obstacles, and the storage location.

    In (\cite{Yousefi2022}), we proposed a human-inspired planning algorithm using a concept we called the \textit{Envelope of Manipulation} $E^M$, which is a curve connecting \textit{key points} $E_i$ (see Figure \ref{fig:pp_EM}), these assembled in a set $\mathcal{E}$. Taking the perspective of a human operator based on our observations in the field, we identified two high-level options in the options space $\mathcal{O}$: 1) $O_{1} \in \mathcal{O}$ that encapsulated the motion along the envelope between two cells, and 2) $O_{2} \in \mathcal{O}$ which encodes the operations inside each cell. The operator may group several objects into a \textit{cluster} in order to cut and grab several trees together, before moving on. The envelope $E^M$ can take any shape; however, based on our field observations, it is well approximated by a circular arc. It is thus possible to encode a sequence of operations as a sequence of the two aforementioned options. The problem of robot planning, therefore, turns into optimizing this sequence of options. The overall task of robot planning includes a hierarchy of subtasks, namely, envelope options (i.e., $O_1$ and $O_2$), and moving the arm or crane along the specified trajectories (MA). We will discuss these further in the subsequent sections.        
\section{Shared Autonomy Design for Feller-Buncher Robot} \label{sec:shared_FB}
    \subsection{Hierarchical Robot Planning}
     \label{sec:path_planning_robot}
        
        \begin{figure}[!tbh]
            \centering
            \includegraphics[width=12cm]{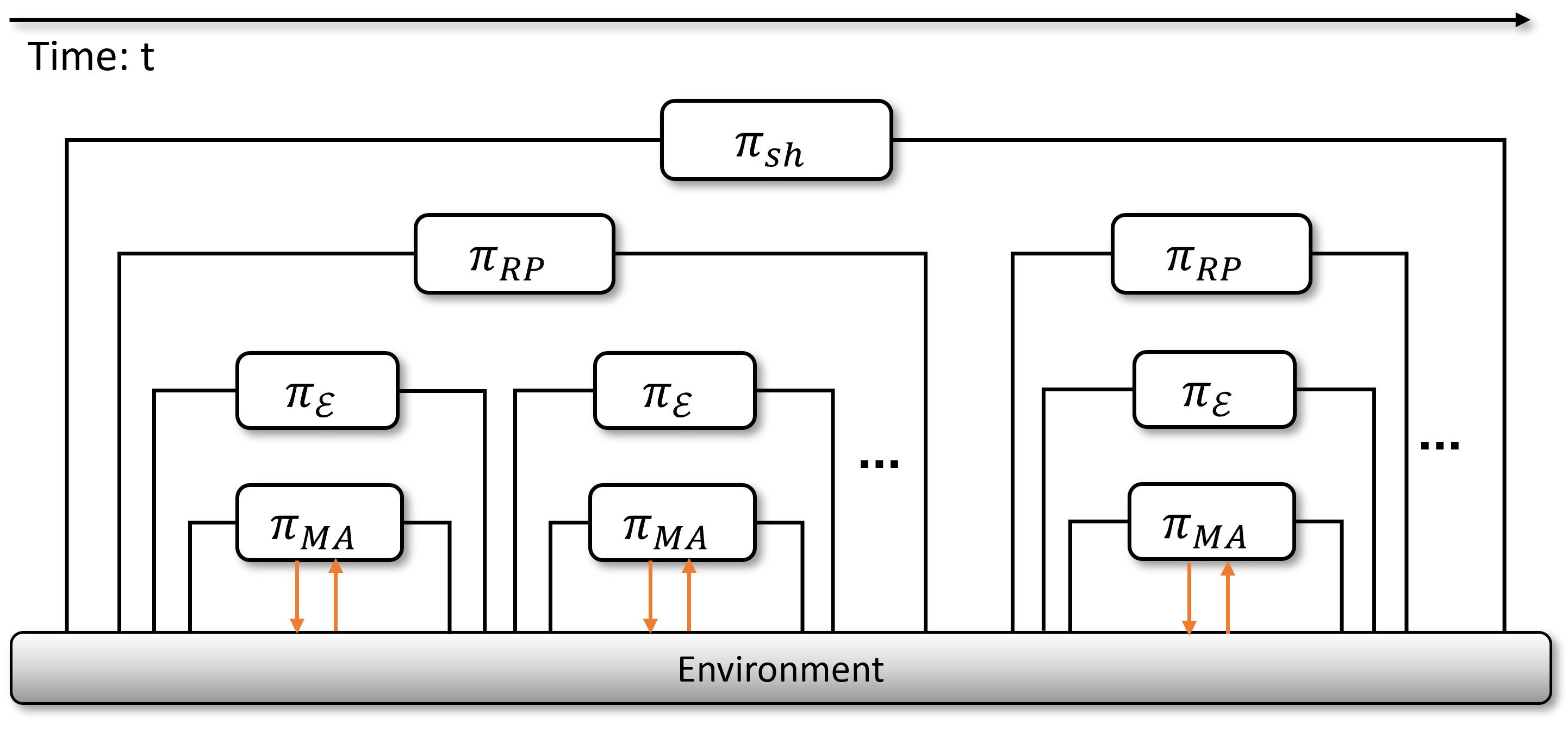}
            \caption{Temporal abstraction of hierarchical shared autonomy policies. This figure shows how \textit{Shared Policy} $\pi_{sh}$ is configured as a higher level policy on top of $\pi_{RP}$ in the general spatiotemporal scheme. This is analogous to Figure \ref{fig:RP_gen_spatiotemporal}, specialized to the feller-buncher robot/machine related tasks.}
            \label{fig:hmdp_temporalabstraction}
        \end{figure}
        \begin{table}[h]
            \centering
            \caption{Break down of robot planning tasks into a hierarchical planning and execution scheme with different levels}
            \begin{tabular}{p{1cm} |p{5.5cm} l}
                \hline
                policy level &  description\\
                \hline
                \hline
               $\pi_{RP}$  &  Overarching policy to plan a robot motion in a task-oriented manner\\
               \hline
               $\pi_{\mathcal{E}}$  &  Policy encapsulated in Envelope of Manipulation ($E^M$) actions, an instance of $\pi_{MHL}$.\\
               \hline
               $\pi_{MA}$  &  Policy to Move Arm (MA), an instance of $\pi_{MLL}$.\\
               \hline
            \end{tabular}
            \label{tab:hierarchical_policies}
        \end{table}
        
         Following \S\ref{sec:pp_gen}, we break down the tasks into a hierarchical planning and execution scheme with the levels listed in Table \ref{tab:hierarchical_policies}, where $\pi_{\mathcal{E}}$ is an instance of $\pi_{MHL}$, and $\pi_{MA}$ is an instance of $\pi_{MLL}$. As noted in \S \ref{sec:application}, in the current scheme of operations, a human operator uses the arm to manipulate (e.g., cut, grab, and deposit) the objects in the operational region of  a particular base location. As listed in Table \ref{tab:hierarchical_policies}, we break down the relevant tasks into a hierarchical planning and execution scheme, with three levels defined as follows: 
        
        \subsubsection{$\pi_{RP}$: Overarching policy to plan a robot motion in a task-oriented manner}\label{sec:CTR}
        
            This is the global or master policy which includes the policies of the lower levels and collects the corresponding rewards; this policy is executed once per \textit{region}.
        
        \subsubsection{$\pi_{\mathcal{E}}$: Policy for Envelope of Manipulation ($E^M$)} \label{sec:E}

            The definition of this level in our hierarchy was motivated by our observations of expert operators: they first implicitly carry out a clustering of trees by grouping subsets of objects into clusters around the machine and subsequently, interact with the objects in clusters. Each cluster can include multiple objects, and span across one or more cells. We designate each \textit{cell} with a \textit{key point} $E_i$ and we have a set of key points $\mathcal{E}$, defined as $\mathcal{E}=\{E_0,E_1,...,E_n\}$, where $E_0$ and $E_n$ correspond to the initial end-effector location at the start of operation and the storage point, respectively. The planning problem is, in fact, a sequential decision making problem to optimally sequence options $O_1$ and $O_2$ and how to group the objects next to a key point as cluster(s).
            
        \subsubsection{$\pi_{MA}$: Policy to Move Arm (MA)} \label{sec:MA}

            This is the lowest-level policy in our hierarchy, and it directly interacts with the environment. This policy takes the destination key point as its goal, plans a smooth trajectory for the end-effector to reach it, and executes the motion of the arm. Standard robotic tools can be employed to execute these subtasks. Although we do not design this policy directly in the present implementation, we include the reward terms related to it in robot planning.

        To find the optimal policy for Robot Planning, $\pi^*_{RP}$, we build on  our previous proof-of-concept formulation in (\cite{Yousefi2022}), combined with the background provided in \S \ref{sec:application}. For that purpose, we construct the more general, compared to (\cite{Yousefi2022}),  Markov Decision Process (MDP) framework, $MDP^{RP}$, as follows:
        
        \noindent \textbf{The Environment or The World}: We have tailored a commonly employed grid world to our specific robotic application, for a generalizable MDP backbone. As schematically shown in Figure \ref{fig:pp_EM}, this environment is comprised of $n_c \times n_r$  cells, circumferentially arranged around the robot base; $n_c$ and $n_r$ are the circular and radial dimensions of the grid, respectively, these chosen  to accommodate the desired resolution.  An important element of environment definition which directly affects $\pi_{RP}$ is how to handle the objects surrounding the robot. These objects, depending on the state of the system, can be either obstacles or subgoals, and this categorization changes dynamically. As illustrated in Figure \ref{fig:rl_env_spaces}, we first construct the relevant \textit{Spaces} as follows:
            \begin{figure}[!tbh]
                \centering
                \includegraphics[width=4cm]{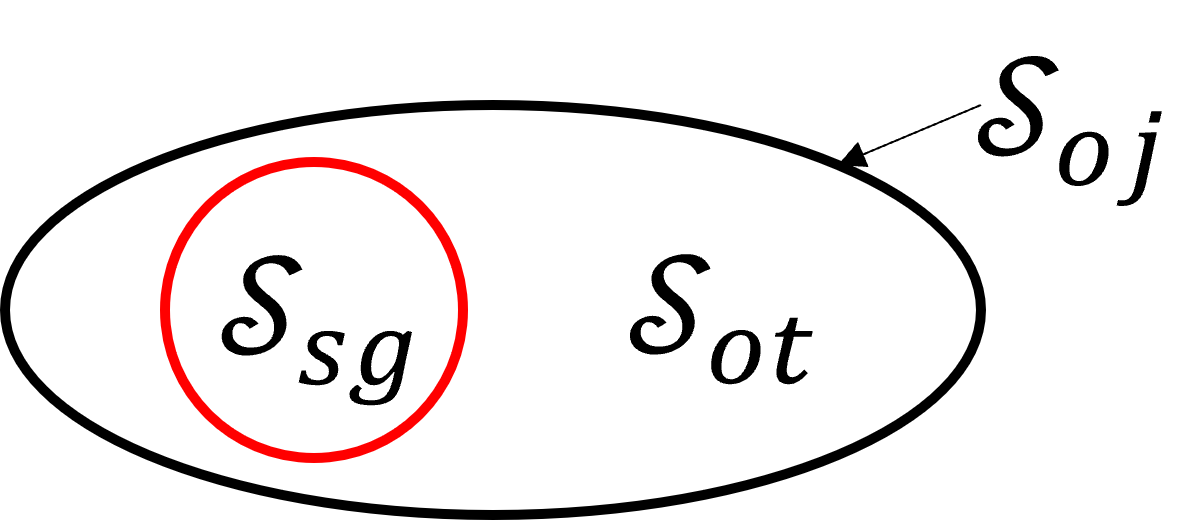}
                \caption{Objects $\mathcal{S}_{oj}$, Obstacles $\mathcal{S}_{ot}$, subgoals $\mathcal{S}_{sg}$ space characterization for the environment in $\pi_{RP}$.}
                \label{fig:rl_env_spaces}
            \end{figure}
            
            \begin{itemize}
                \item Objects space  $\mathcal{S}_{oj}$: includes all objects. 
                \item Subgoals space $\mathcal{S}_{sg}$: a subset of $\mathcal{S}_{oj}$ and it includes all objects accessible to  the robot (and hence, not blocked by other objects from robot's reach). The augmented Subgoal space, $\hat{\mathcal{S}}_{sg}$, is constructed by adding the storage point, as  conditioned on the capacity for maneuverability  $CfM_{ee}$, with the following logic:
                \begin{itemize}
                    \item if $CfM_{ee}=0$, $\hat{\mathcal{S}}_{sg} = \{E_n\}$
                    \item if $0 < CfM_{ee} < 1$, $\hat{\mathcal{S}}_{sg} = \{\mathcal{S}_{sg}, E_n\}$
                    \item if $CfM_{ee}=1$, $\hat{\mathcal{S}}_{sg} = \{\mathcal{S}_{sg}\}$.
                \end{itemize}
                \item Obstacles space $\mathcal{S}_{ot}$: defined by negating the augmented Subgoal space from the Objects space. Therefore, the three object spaces are related through:
                \begin{equation}
                    \mathcal{S}_{oj} = \mathcal{S}_{sg} \cup \mathcal{S}_{ot}
                \end{equation}
            \end{itemize}
            \begin{figure}[!tbh]
                \centering
                \includegraphics[width=5cm]{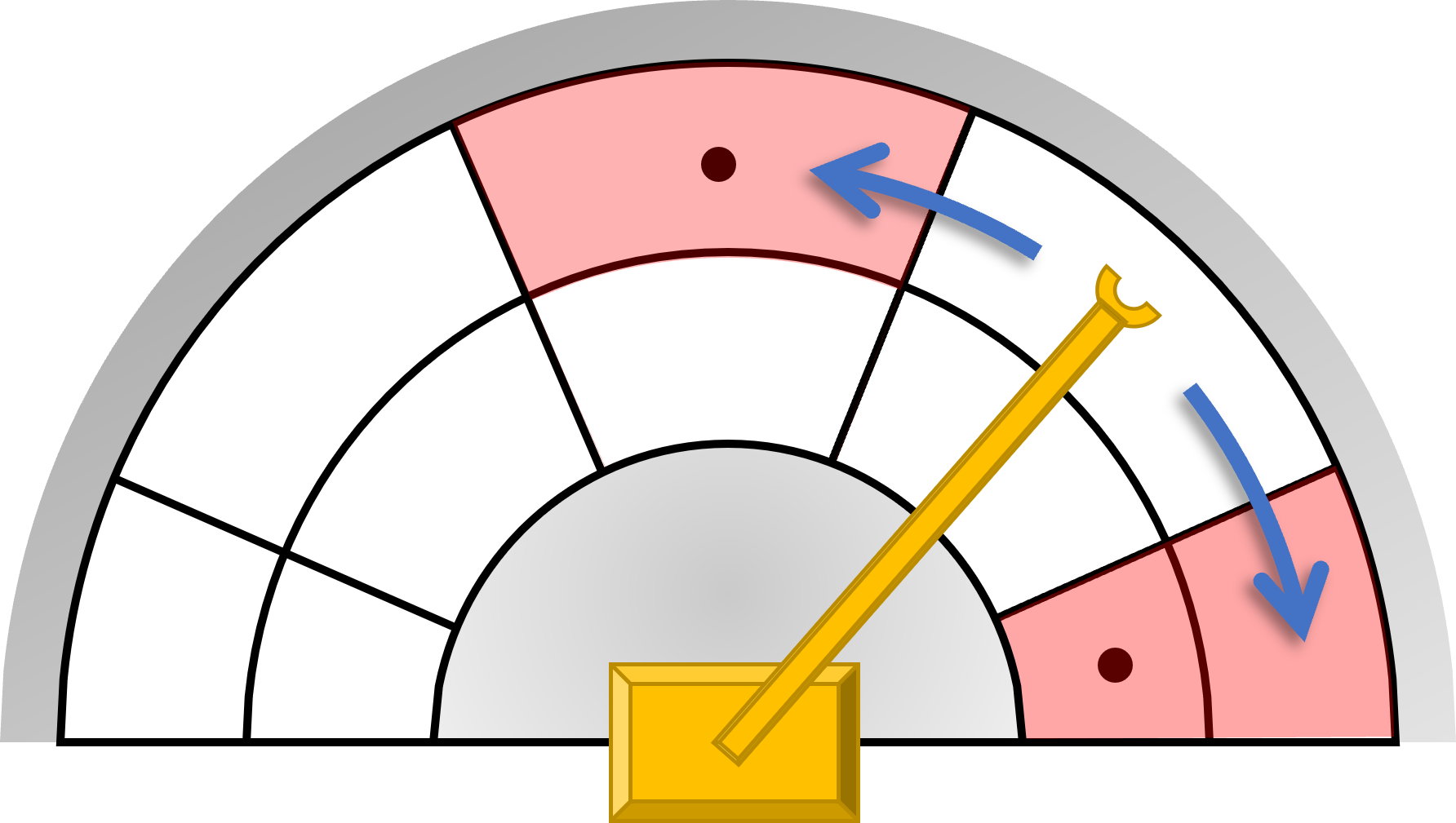}
                \caption{Illustration of obstacles obstructing the movement of the arm in different directions from a cell in the grid with the arm colored in amber. The pink cells designate "obstacle" cells.}
                \label{fig:rl_env_shadowObs}
            \end{figure}
        \noindent \textbf{Constraints}    
             In general, the workspace of the robot is limited by (a) boundaries of the grid, which are in turn defined by the minimum and maximum values of the reachability of the robot, $CfM_{arm}$, and (b) obstacles located next to the end-effector as well as those obstructing its arm movement, as depicted in Figure \ref{fig:rl_env_shadowObs}. This environment, therefore, includes the following constraints built-in and dynamically updated:
            \begin{itemize}
                \item Robot workspace and related constraints, such as the capacity for maneuverability of the arm and the end-effector, i.e., $CfM_{arm}$ and $CfM_{ee}$, respectively. This can be extended to stability related constraints, as well.
                \item Path planning related constraints, such as obstacles.
            \end{itemize}
            
            With the definition of the world, the categorization of object spaces, and the constraints, we are now ready to define the main MDP elements which in turn encode the path planning problem with obstacle avoidance, as discussed earlier.
            \begin{figure}[!h]
                \centering
                \includegraphics[width=7cm]{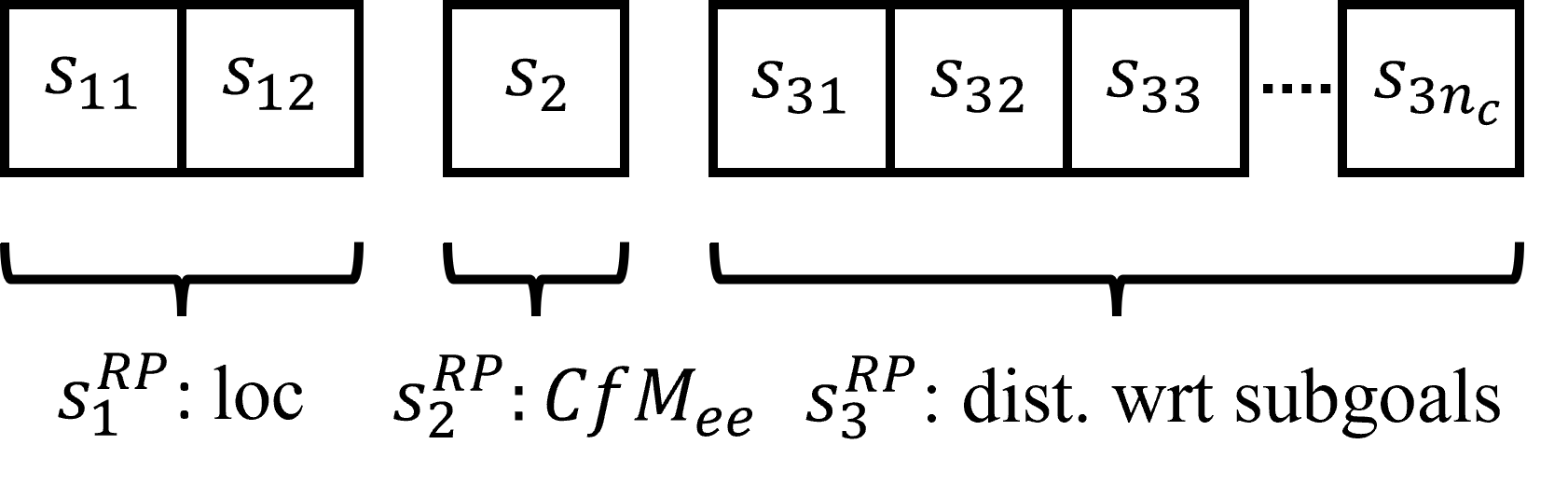}
                \caption{State definition for $\pi_{RP}$}
                \label{fig:level_em_statedef}
            \end{figure}
            
            \noindent \textbf{State (or Observation) Space}. As shown in Figure \ref{fig:level_em_statedef}, the observation space is a discrete space with three types of observations:
            \begin{enumerate}
                \item $s^{RP}_1$: discrete 2D position of the robot end-effector, [angular position, radial position], in the range $0,..,n_c-1$ and $0,...,n_r-1$, respectively,
                \item $s^{RP}_2$: payload indicator $0, ..., p_{max}$;  it is related to $ CfM_{ee}$ through  ($CfM_{ee}=(p_{max}-s_2)/p_{max}$),  where $p_{max}$ is the maximum number of trees/objects that the end-effector is able to carry.
                \item $s^{RP}_3$: contains the circular distance from the current cluster/key point to all subgoals with respect to the robot end-effector in the CCW direction. If no subgoal is  present at a location, -1 is returned. Therefore,  $s^{RP}_3$ effectively augments the state of the robot (comprised of $s^{RP}_1$ and $s^{RP}_2$) with the information related to the subgoals space. It is, in fact, the variable $z_0$ introduced earlier since it  encodes the goal space information of the task.
                
                Therefore, we denote the state in this level with $s^{RP}$ defined as follows:
                \begin{equation} \label{eq:s_E_def}
                    s^{RP} \triangleq
                    (s^{RP}_1,~s^{RP}_2,~s^{RP}_3).
                \end{equation}
            \end{enumerate}
            Together with the spaces defined above, a state $s^{RP}$ encodes three features depending on the scenario:
            \begin{enumerate}
                \item obstacle, if $s^{RP} \in \mathcal{S}_{ot}$,
                \item sub-goal, if $s^{RP} \in \hat{\mathcal{S}}_{sg}$. If the agent is \textit{done} with the operation overall, reaching the storage point is the \textit{final goal} and the episode is \textit{done}. 
                \item normal, otherwise.
            \end{enumerate}
            
            \noindent \textbf{Action Space}. The action space is discrete, consisting of four actions: left, right, front, and back, encoded by 0, 1, 2, and 3, respectively. Note that the directions of these actions are defined with respect to each cell, and not in an absolute sense.
            
            \noindent \textbf{Rewards}. Rewards are defined as follows: 
                \begin{itemize}
                    \item $R^{RP}_1 = -2$: All transitions except the transition to the ”sub-goal” or ”goal” state,
                    \item $R^{RP}_2 = 20n_{cut}~\text{or}~20n_{store}$: Transition to one of the ”sub-goal” states for the cases of cutting or storing,
                    \item $R^{RP}_3 = 400$: Transition to the ”goal” state. This ends an episode and resets the environment,
                    \item $R^{RP}_4 = -20$: Collision with an obstacle,
                    \item $R^{RP}_5 = -20$: Out of boundaries action,
                    \item $R^{RP}_6 = -5s^{RP}_2$: Cost of carrying an object,
                    \item $R^{RP}_7 = -400$: Getting trapped. This also ends an episode and resets the environment.   
                \end{itemize}
            Note that the combined value of the above reward elements forms $R_1$ in \eqref{eq:shared_reward}.  
                
            \noindent \textbf{Policy}. With the architecture shown in Figure \ref{fig:h-shared-RL-arm}, we denote the policy for this level with $\pi_{RP} = \pi_{RP}(o^{RP,A}|s^{RP})$.

            It is worth noting that to implement this world efficiently, we have created a custom OpenAI Gym (\cite{brockman2016openai}) environment, and we call it ``adaptive\_grid\_v0".

    \subsection{Shared Autonomy Setting}
        As shown in Figures \ref{fig:sharedRL_hierarchical_comp} and \ref{fig:hmdp_temporalabstraction}, we define the autonomous agent policy as the highest level policy called \textit{Shared Policy}, $\pi_{sh}$, and model it as an instance of MDP with the following elements:
            
        \noindent \textbf{The Environment or The World}: We designed a higher level environment for the tasks of shared autonomy for a generalizable MDP backbone, the attributes of which are defined shortly. In particular, we have created a second custom OpenAI Gym environment, called ``assist\_AI\_v0", which  directly communicates with the lower level environment, adaptive\_grid\_v0, and acts as a master agent with a master policy incorporating the assistance and/or autonomy protocols.
         
        \noindent \textbf{State Space}. This is defined based on state space of the $MDP^{RP}$, but expands it to include $z_1$ and $a^H$. Note that $a^H$ is recorded as -1 if no action is taken by the human agent to accommodate such instances. 

        \noindent \textbf{Action Space}. This is defined similar to that of $MDP^{RP}$, and is comprised of four actions: left, right, up, and down, encoded by 0, 1, 2, and 3, respectively.

        \noindent \textbf{Policy}. The policy considering the model presented in Figure \ref{fig:sharedRL_layer_gen} and discussions in \S \ref{sec:autnomous_agent}.

        \noindent \textbf{Rewards}. Rewards, as discussed earlier, take the form of $\mathbi{c}^\top\mathbi{R}$ in \eqref{eq:shared_reward}.
        
        From another perspective, since the autonomous agent is  human-inspired  by design, with a shared autonomy mindset, the autonomous actions are comprehensible to  the human agent and vice versa. Hence, this shared mental model (\cite{Chris2007,Yousefi_MScThesis2018}) is not only necessary for proper collaboration but more importantly provides a road map to the design of any modern framework involving humans and (semi-)autonomous agents. Our approach has the capability to correct and help train a novice human operator in a safe and efficient manner. On the other hand, with the designed hierarchical learning and planning algorithms, full autonomy is also achievable.


\section{Numerical Results \& Experiments} \label{sec:results}
    
    We present numerical  results in four parts, progressively adding layers of complexity, in a similar order to the material discussed in \S \ref{sec:elements}. The four sets of results also illustrate how we build up and test our shared autonomy framework in the following stages:
        \begin{itemize}
            \item \textbf{Stage I}: Pre-Training\textemdash this stage produces an autonomous agent trained by using deep RL (RL) techniques. This model will be considered as the baseline to which the behavior of a human operator  will be compared. The results for this stage are presented in \S \ref{sec:results_pp}. Since our shared autonomy framework is capable of full autonomy by design, i.e., highest level of autonomy, the overarching goal in this stage is to showcase such capability in training and testing of an autonomous agent given the inherent stochasticity of the environment as well as the challenges of the application.
            \item \textbf{Stage II}: Manual\textemdash we let the human take full control authority over the robot. The advantage of a shared autonomy framework is in effectively incorporating human operator in the loop to learn from and, in general, to switch control if needs be. The results of this stage, presented in \S \ref{sec:results_for_human_analysis} are to set this important capability in place and provide the algorithm with necessary data for a human-tuned framework using the formulation presented in \S \ref{sec:human_analysis}.
            \item \textbf{Stage III}: Shared-Training\textemdash we train the shared autonomy policy according to our proposed model, also by using deep RL techniques. Our approach in this stage is looking into certain challenging \textit{scenarios} in training a shared autonomy agent with expert and noisy humans and to see the effects of different components of our architecture. The results for this stage are presented in \S \ref{sec:results_for_sharedAutonomy}.
            \item \textbf{Stage IV}: Shared-Testing\textemdash we test the trained model with expert human for a variety of cases. Finally, in this stage, we interface the trained shared autonomy agent with humans with different levels of noise and analyze its performance. The results for this stage are presented in \S \ref{sec:results_for_sharedAutonomy_part2}.
        \end{itemize}
    
    \begin{figure}[!tbh]
        \centering
        \includegraphics[width=10cm]{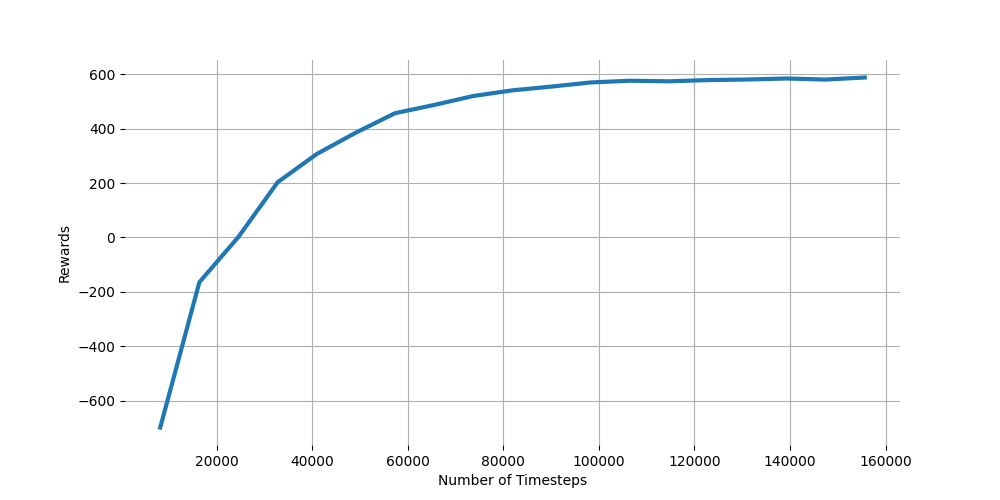}
        \caption{Training of autonomous policy using a deep RL technique, PPO.}
        \label{fig:rl_deepRL_AGv0}
    \end{figure}
    \begin{figure}[!tbh]
        \centering
        \includegraphics[width=6cm]{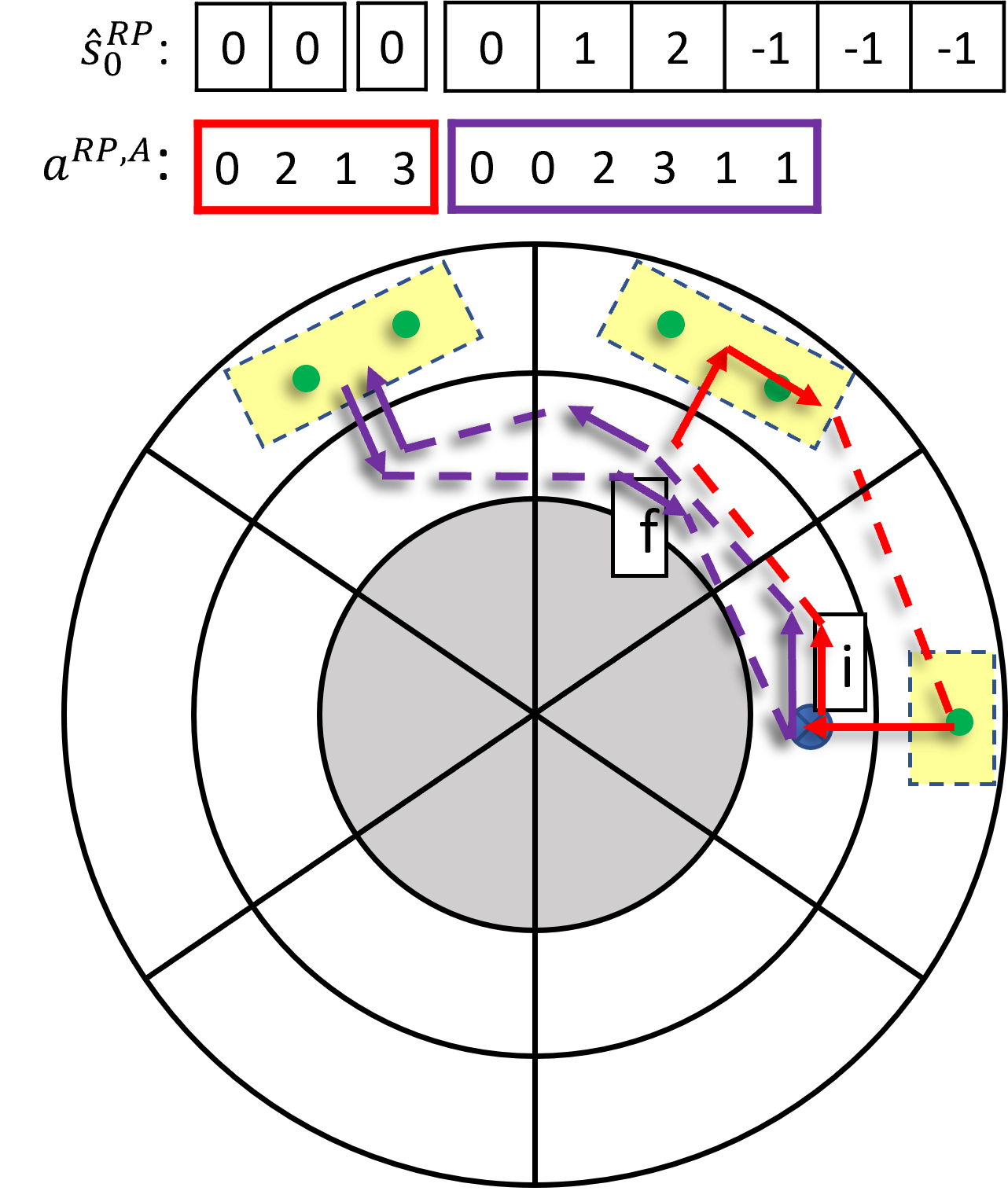}
        \caption{Pre-trained agent Example 1.  Green circles are objects (trees), yellow rectangles are clusters,  blue circle is the initial agent location, solid arrows indicate actions, dashed lines conceptualize the path (simplified to straight lines for ease of illustration). The sequence starts at ``i" and continues  until the last action ``f". $CFM_{ee}$ is set to 4. For better readability, we use different colors for  the two  sub-sequences ending at a  storage location (red, purple, in that order). The storage is also located at ``i".}
        \label{fig:exa1_EM_fig}
    \end{figure}
    \begin{figure}
        \centering
        \includegraphics[width=9cm]{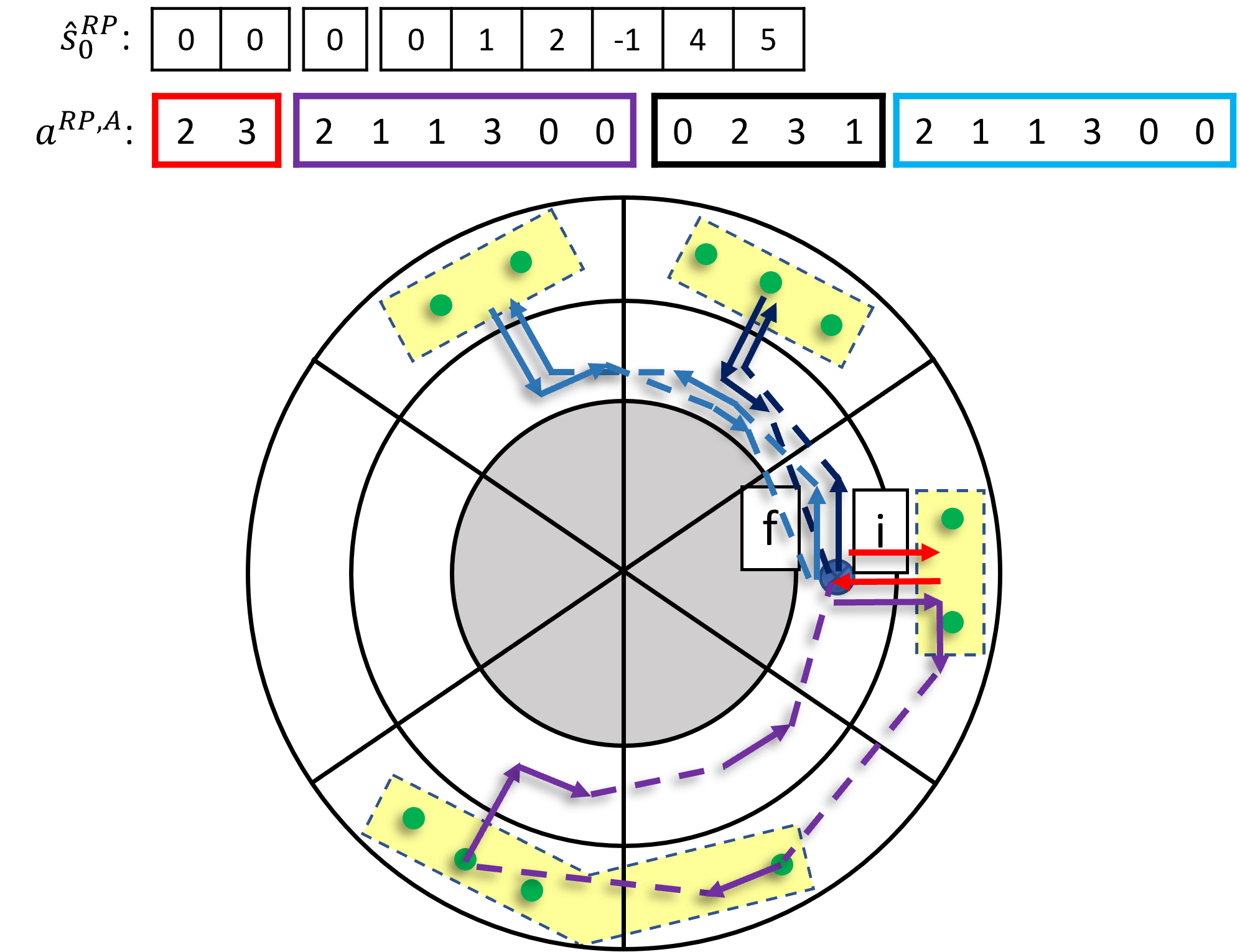}
        \caption{Pre-trained agent Example 2. Green circles are objects (trees), yellow rectangles are clusters,  blue circle is the initial agent location, solid arrows indicate actions, dashed lines conceptualize the path (simplified to straight lines for ease of illustration). The sequence starts at ``i" and continues  until the last action ``f". $CFM_{ee}$ is set to 4. For better readability, we use different  colors for the four sub-sequences, each ending in storage location (red, purple, black, and blue, in that order). The storage is also located at ``i".}
        \label{fig:exa2_EM_fig}
    \end{figure}
    \subsection{Results for Stage I, Pre-Training} \label{sec:results_pp}\label{sec:results_for_robot_planning}
        Here, we showcase the training of a fully autonomous robot. In the context of our shared autonomy framework, this will represent a pre-trained agent, to be used as the baseline for  computing the human agent's error.        
        
        With the adaptive\_radial\_grid\_v0 environment described earlier, we use Stable-Baselines3 library (\cite{stable-baselines3}) to train a deep RL policy. During the training process, we sample the objects in the environment from a Gaussian distribution in order to account for different possible variations of the object spaces. More specifically, we draw samples from a truncated Gaussian distribution in interval $[0~4]$ with the mean and standard deviation of 2 and 1, respectively. Accordingly, our formulation is  \textit{uncertainty-aware}. We  employ the Proximal Policy Optimization (PPO) algorithm (\cite{Schulman2017}) notably with a batch size of 32, learning rate of $1\times 10^{-3}$, and $\gamma$ of 0.99 for a multilayer-perceptron (MLP) policy with 2 layers of 64 nodes. Figure \ref{fig:rl_deepRL_AGv0} shows the performance of the training process. We also provide two examples of output sequences for two objects configurations: Figure \ref{fig:exa1_EM_fig} for a relatively sparse scenario, and Figure \ref{fig:exa2_EM_fig} for a relatively crowded scenario. The figures include information on the initial state and the output action sequences of the policy. We observe that these are logical and intuitive. 
       
    \subsection{Results for Stage II, Manual} \label{sec:results_for_human_analysis}
        
        Here, we present the results supporting the encoding of the human agent's internal/latent variable $z_1$. We build up a shared autonomy platform that, at its core, is comprised of our hierarchical MDPs implemented using two Gym environments. We have used Cogment (\cite{cogment}) to enable real-time human-in-the-loop interaction in our platform. Cogment platform is an open source framework built on a micro-service architecture for running different kinds of RL, multi-agent RL and human-in-the-loop learning applications. 

\begin{figure}
            \centering
            \includegraphics[width=10cm]{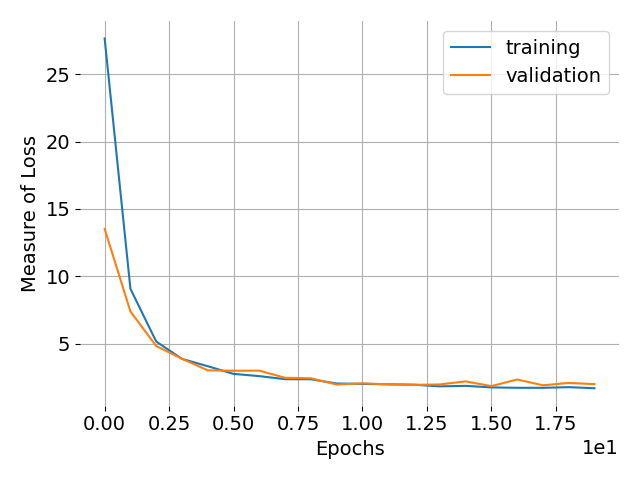}
            \caption{Training vs Validation for cVAE model.}
            \label{fig:tr_val_cVAE}
        \end{figure}

\begin{figure}[!tbh]
            \centering
            \includegraphics[width=6cm]{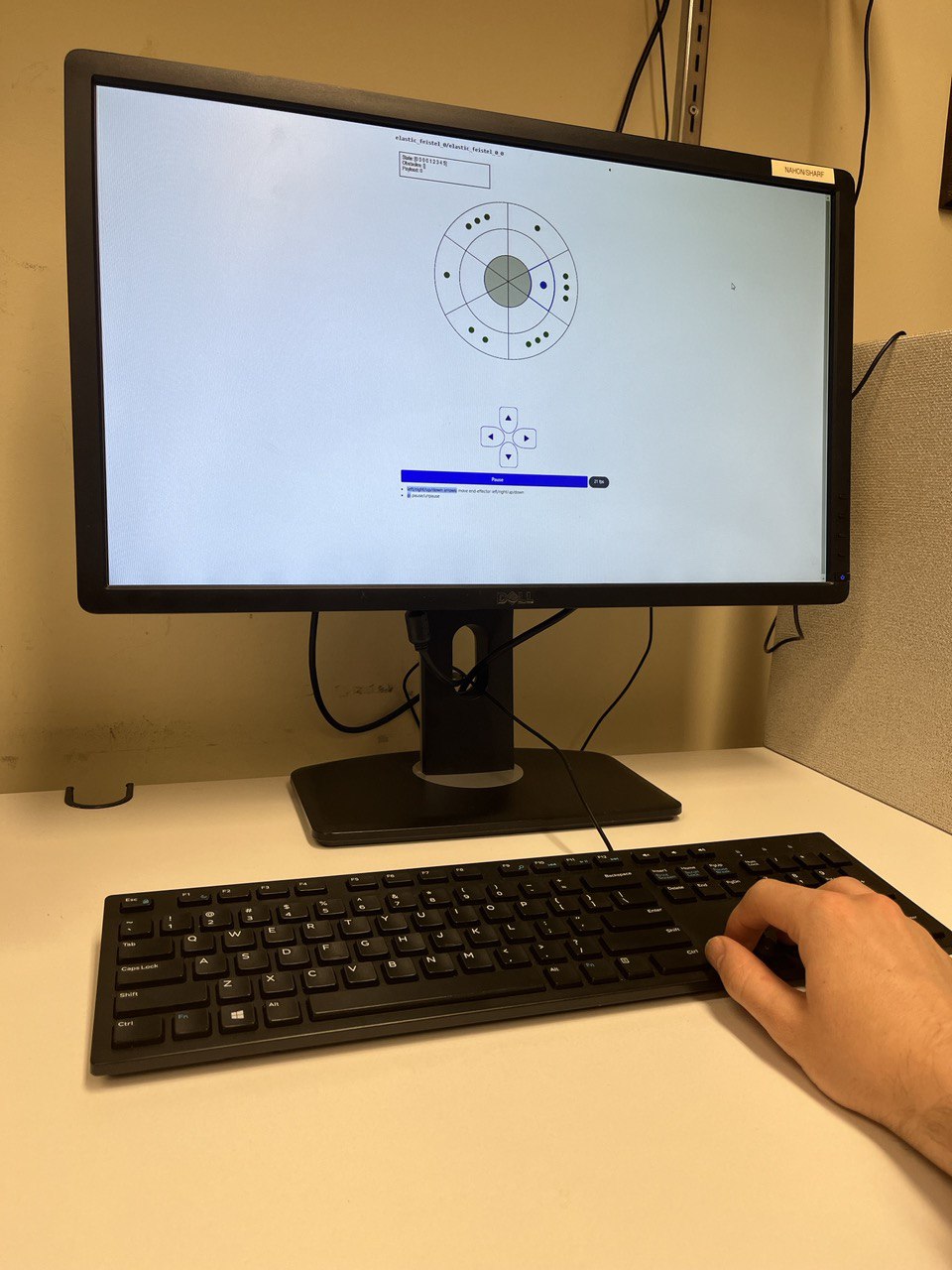}
            \caption{Our setup to conduct human-in-the-loop tests. We use a visualization layer to show the environment to the user, who can move with 4 basic inputs in 4 directions, a reflection of the inputs defined in our discrete robot planning MDP introduced in \S \ref{sec:shared_FB}.}
            \label{fig:setup}
        \end{figure}

        During a test, human user is presented with a random initialization of the environment, an example of which is shown Figure \ref{fig:exa3_EM_fig}. The four basic discrete inputs, introduced in \S \ref{sec:path_planning_robot}, are mapped to four direction buttons on a regular keyboard. Figure \ref{fig:setup} shows our setup for a test. We record the actual human data using similar object randomization and environment configurations as in Stage I. It is also important to note that an explicit goal inference is not feasible in our set-up except for a myopic one (\cite{dragan2013policy}), which assumes that the intended goal is the closest of goal space points. This, in essence, is how we defined variable $z_0$ in this work that encodes the angular distance to the nearby goals. Following the formulation presented in \S \ref{sec:human_analysis}, we train our auto-encoder for a 5D latent variable $z_1$ using 2 history steps ($n_h=2$) using 40 recorded episodes or trials of a human user interacting with our setup. It is worth noting that we randomly divide the dataset with a ratio of 0.7. Notably, the learning rate and batch size are $5\times10^{-4}$ and 5, respectively. To compute the input error of \eqref{eq:eH}, we use the pre-trained model from Stage I as the surrogate optimal model. From a practical point of view, we used one-hot transformation for our discrete variables, such as the state $s^{RP}$, and introduced white noise for better training. Figure \ref{fig:tr_val_cVAE} shows the training and validation process of our cVAE model.
        
    \subsection{Results for Stage III, Shared-Training} \label{sec:results_for_sharedAutonomy}

        Building on Stages I and II, Stage III is the next step in setting up our platform. In this work, we used deep RL to train the shared policies using PPO algorithm (\cite{Schulman2017}), similar to the pre-trained model. We consider three scenarios of training with different human skill levels and different shared autonomy settings. Each of these scenarios might include one or more hypotheses, i.e., a research question(s), followed by presentation of the results and assessing them. It is worth noting that in evaluating the results of the training process, we use the following two measures:
        \begin{itemize}
            \item Reward per time-step: the rewards with respect to the training (simulator) time-step. In all training cases, we train the policy for a total number of $N_{tr}=5\times10^{5}$ time-steps, which is the horizontal axis in all rewards-related plots. Note that we use subscript $tr$ for training related variables.

            \item Sample Processing Rate (SPR): the number of processed samples forward and backward per second. In shared autonomy, the speed of the platform is of crucial importance, since it needs to train an agent for a shared autonomy while the task progresses with human in the loop. SPR is defined as follows:
            \begin{equation} \label{eq:SPR}
                SPR(k) = \sum_{j=0}^{k} n^{j}_{tr} / (t-t_0),
            \end{equation}
            where $t$ and $t_0$ are the current and initial wall-time in seconds, $n^{k}_{tr}$ is the number of training samples processed at time-step $k$. This also includes the time required by the stochastic gradient decent. We use Adam as our optimizer (\cite{kingma2017adam}). Batch size is 64 with learning rate of $1\times10^{-4}$. 
        \end{itemize}

        Moreover, we have two expertise levels for the human agent in our trials:
        \begin{itemize}
            \item \textit{Expert Human}: A human agent who is familiar with our setup.
            \item \textit{Noisy Human}: We deliberately perturb the human's action by adding noise.
        \end{itemize}
        In all cases, if no human action is available, the human agent is considered to be non-collaborative. No action is an action itself. We use the above-mentioned metrics to investigate: (a) whether or not a shared autonomy agent can be trained under different human expertise levels and the degree of collaborativeness given the proposed MDP structure, (b) to what extent the human-tuned variable $z_1$ affects the training process, (c) how rewards terms and their coefficients in \eqref{eq:shared_reward} affect the training process, (d) how a trained model performs if interfaced with humans of different expertise level.
        
        \vspace{.1cm}
        
        \noindent \textbf{Scenario 1: Training with Expert Human}
        
        We begin the presentation of shared autonomy results  by demonstrating  the training process for expert human, defined above, which includes the trials data collected from a human agent familiar with our setup. Arguably, no human-in-the-loop test can cover the complete state space, and therefore, we will treat the human agent at the unseen states also as a non-cooperative agent who takes no action. In this scenario, we choose equal weights for $R_1$ and $R_2$ and set $\mathbi{c} = [10, 10]$.
        \begin{hypothesis} \label{hyp:hypothesis1}
            A shared autonomy policy can be trained using our formulation for an expert human, as defined before, who is alternating between cooperative and non-cooperative, with the inherent stochasticity of the environment.
        \end{hypothesis}
        \begin{figure}[!tbh]
            \centering
            \includegraphics[width = 9cm]{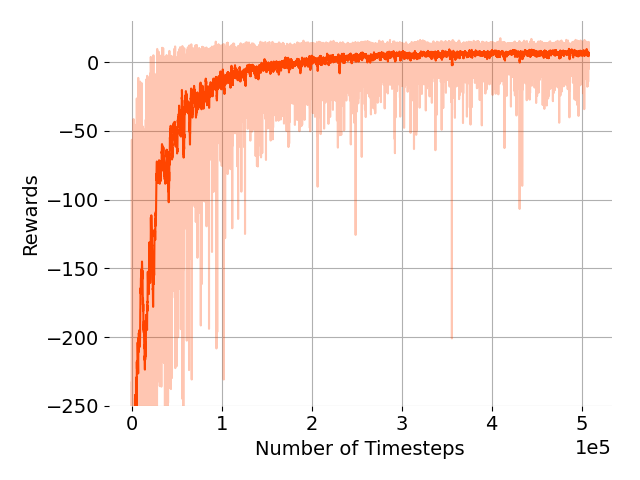}
            \caption{Training process for scenario 1: Expert human. The darker red is the smoothed, average reward with a window of 50 time-steps. The lighter red shows the raw training rewards.}
            \label{fig:opt_human}
        \end{figure}
        Figure \ref{fig:opt_human} shows the training process, confirming the success of our algorithm in training a shared autonomy agent detailed in Hypothesis \ref{hyp:hypothesis1}. Further tests with the trained model will be provided in \S \ref{sec:results_for_sharedAutonomy_part2}. 
        
        \begin{hypothesis} \label{hyp:hypothesis2}
            Under the conditions outlined in Hypothesis \ref{hyp:hypothesis1}, we hypothesize that the human-tuned variable $z_1$ results in a more efficient training process.
        \end{hypothesis}
        \begin{figure}[!tbh]
            \centering
            \includegraphics[width = 8cm]{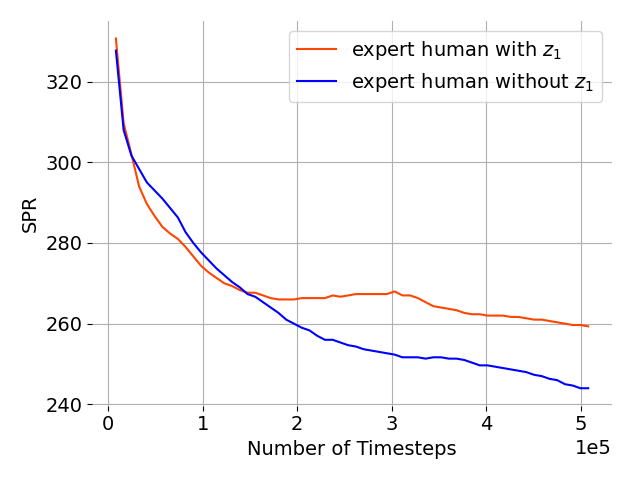}
            \caption{Sample processing rate (SPR) for Scenario 1: Expert human with and without $z_1$.}
            \label{fig:opt_human_fps_comparison}
        \end{figure}

        Figure \ref{fig:opt_human_fps_comparison} shows how sample processing rate (SPR), defined in \eqref{eq:SPR}, changes with respect to training time-step for the expert human \textit{with} and \textit{without} $z_1$. In the case of without $z_1$, we still pass the signal through the $z_1$ model but zero it out before feeding to the algorithm in order to as much as possible isolate the mere effect of $z_1$. In the early stages of the training time-steps, i.e., $ts\leq 1.6 \times 10^5$, which is the highly oscillatory stage, we do not see noticeable difference between the performances of the two cases. However, as the training progresses, the effect of $z_1$ is evident, which results in more efficient performance. This result partially confirms Hypothesis \ref{hyp:hypothesis2}. Intuitively, $z_1$ is effective when the policy is getting closer to the convergence. Moreover, it can be concluded that $z_1$ contributes positively in our shared autonomy framework for an expert human.

        \vspace{.1cm}
        \noindent\textbf{Scenario 2: Training with Noisy Human}
        
        In this scenario, we deliberately perturb the human action by adding noise to their action, that results in a noisy human, defined above. The process is outlined in Appendix B. This is a challenging scenario for our setup for three reasons: (a) at a conceptual level, a noisy human in shared autonomy setting is, in general, a non-collaborative agent, which makes the task of the autonomous agent that is accommodating them challenging. As noted in \S \ref{sec:intro}, the aspect where \textit{policy shaping} outshines \textit{policy blending} is in this interfacing with a noisy human in the loop, (b) as we are assessing the limits of our setting, we are still using equally weighted rewards with $\mathbi{c} = [10, 10]$. We will drop this constraint later. (c) we keep using the $z_1$ variable that is fine-tuned to the expert human. The latter results in mismatched $z_1$, which makes the process even more challenging.
        \begin{figure}[!tbh]
            \centering
            \includegraphics[width = 8cm]{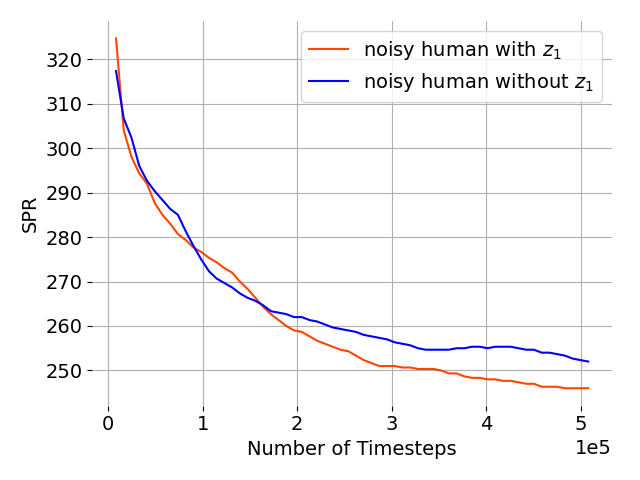}
            \caption{Sample processing rate (SPR) of the training process for scenario 2: Noisy human with and without $z_1$.}
            \label{fig:noisy_human}
        \end{figure}
        \begin{hypothesis} \label{hyp:hyopthesis3}
            For a noisy human with mismatched $z_1$, we expect this variable to negatively affect the training process.
        \end{hypothesis}

        Figure \ref{fig:noisy_human} depicts the sample processing rate (SPR) for the training process with a noisy human for the cases with and without $z_1$. Similar to the previous assessment, this figure partially verifies Hypothesis \ref{hyp:hyopthesis3}. Only towards the later stages of the training that we observe the effects of including $z_1$, which is negatively affecting the performance. Intuitively, this is expected, since $z_1$ is fine-tuned for a different human.
        
        The results of assessment of Hypotheses \ref{hyp:hypothesis2} and \ref{hyp:hyopthesis3} are significant in the sense that they confirm the validity of our architecture design and the considered assumptions. From the behavior of the training processes, one realizes the inherently complex dynamics of a shared autonomy setting with a human in the loop, and how finding and incorporating human's latent variable improves the training process.
        
        Another aspect of our framework is the ability to adjust the coefficients through $\mathbi{c}$. To showcase this, we present results with reduced $c_2$, i.e., reduced contribution of human action in reward function, and use $\mathbi{c} = [10, 5]$ to counter-act the high noise in human actions. 
        
        \begin{hypothesis}\label{hyp:hypothesis4}
            Given a noisy human, reducing the associated coefficient in objective function, i.e., $c_2$, results in improved training process.
        \end{hypothesis}
        
\begin{figure}[!tbh]
            \centering
            \includegraphics[width = 9cm]{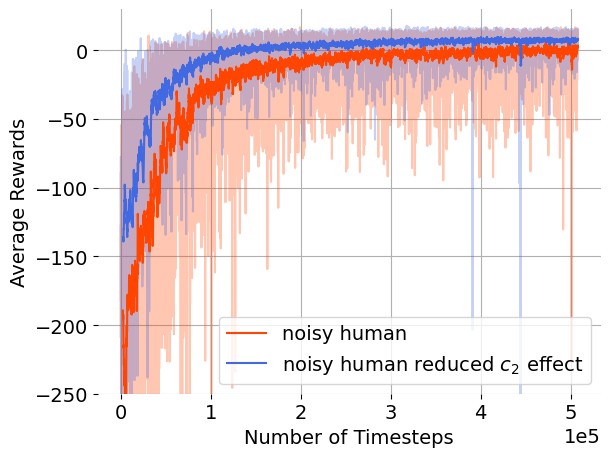}
            \caption{Training process for scenario 2: Noisy human with reduced $c_2$ effect.}
            \label{fig:reduced_c2_effect}
        \end{figure}

        Figure \ref{fig:reduced_c2_effect} compared training performance for the noisy human with and without reduced $c_2$ effect. This figure confirms Hypothesis \ref{hyp:hypothesis4} by showing a much less oscillatory training process and earlier convergence. This result also confirms practical applicability of human-related coefficient $c_2$ as a design variable to control the performance of shared autonomy.
        
        \vspace{.1cm}
        \noindent \textbf{Scenario 3: Training with Override Option}
        
\begin{figure}[!tbh]
            \centering
            \includegraphics[width = 9cm]{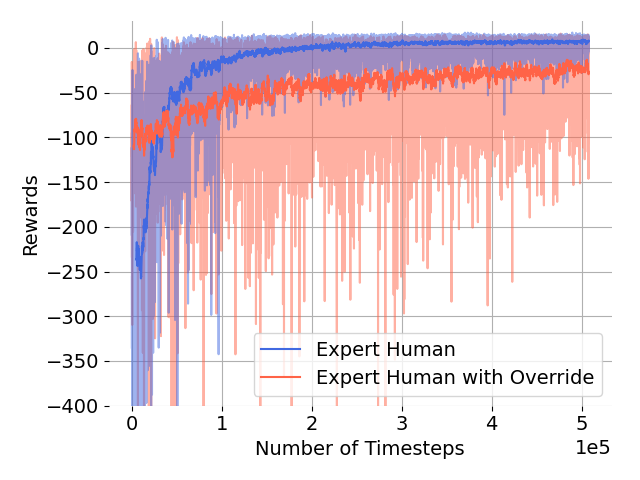}
            \caption{Training performance for Scenario 3: Expert human with and without override option.}
            \label{fig:optimal_vs_override_optimal}
        \end{figure}

        A challenging task for an autonomous agent in shared autonomy is presented when the human agent is given an override option. We tested for the scenario that the human action overrides the autonomous action with a probability of 80\%. This scenario is important from a practical point of view for safety critical applications. The training process is shown in Figure \ref{fig:optimal_vs_override_optimal} which indicates a much more challenging policy update and a sub-optimal policy. However, we maintain the equal weights on $R_1$ and $R_2$. This is another manifestation of the bottleneck in shared autonomy discussed in Scenario 2. That is we are imposing our \textit{designed} definition of a reward function next to what a human \textit{might} consider (and hence, override).

        \subsection{Results for Stage IV, Shared-Testing} \label{sec:results_for_sharedAutonomy_part2}
        \begin{figure}[!tbh]
        \centering
            \includegraphics[width=6cm]{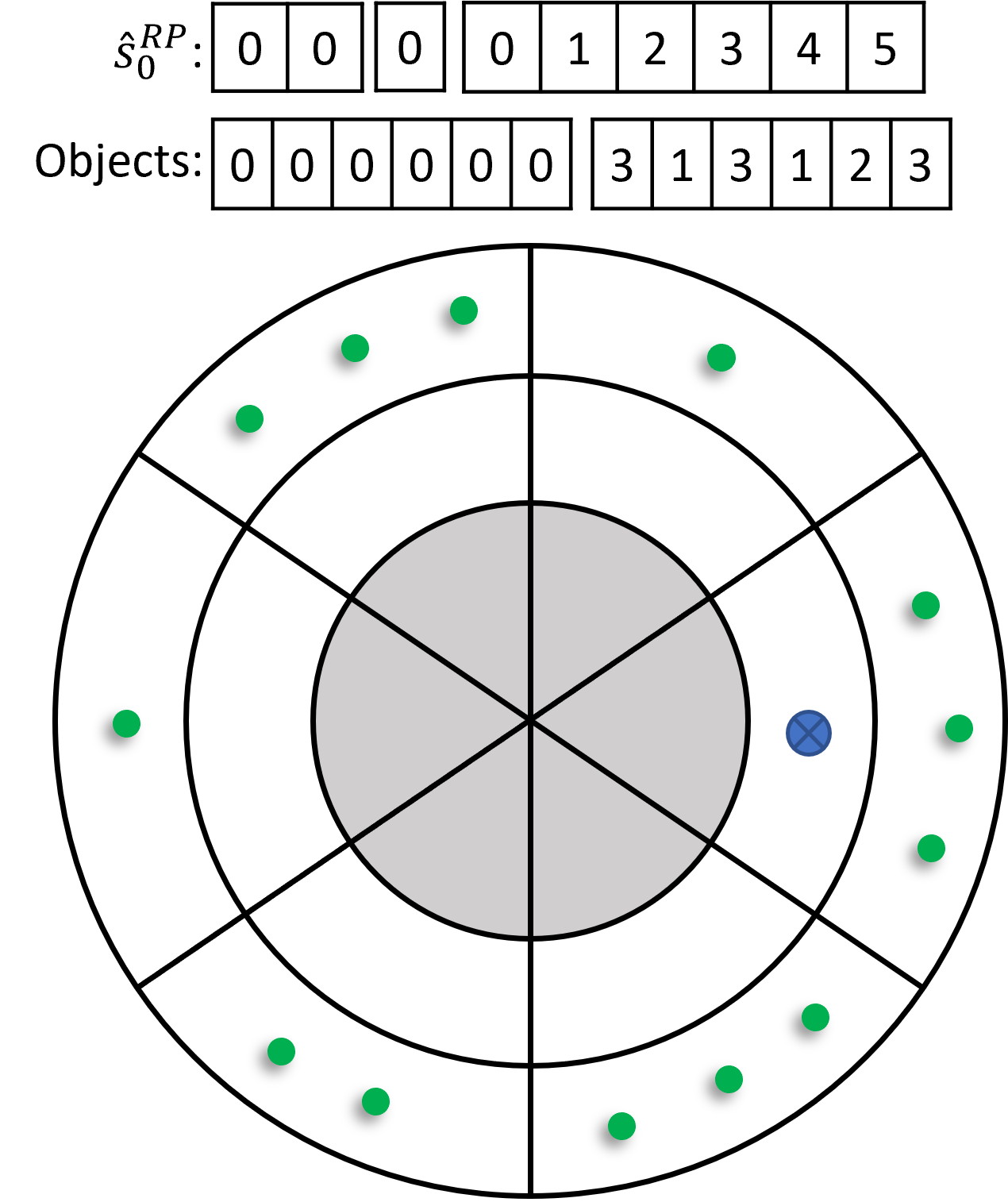}
            \caption{Starting configuration for $\pi_{sh}$ examples}
            \label{fig:exa3_EM_fig}
        \end{figure}
        \begin{table*}[!tbh]
            \caption{Shared autonomy stats for Case 1}
            \centering
            \begin{tabular}{c|c}
                    AA sequence & [1, 2, 0, 3, 2, 3, 0, 0, 2, 1, 1, 3, 2, 1, 1, 0, 0, 3, 1, 1, 1, 2, 3, 0, 0, 0] \\
                    \hline
                    HA sequence &   [1, 2, -1, 1, 3, -1, 0, -1, -1, -1, 2, 1, 2, 3, 2, -1, -1, -1, 1, 1, 2, 0, 1, 0, 1, 0]\\
                    \hline
                    HA Interaction & 18  out of  26, i.e., 69.2\% \\
                    \hline
                    AA followed HA & 8\\
                    \hline
                    Reward & 18.27\\
                \end{tabular}
                \label{tab:results_case1}
        \end{table*}
        \begin{table*}[!tbh]
            \caption{Shared autonomy stats for Case 2}
            \centering
            \begin{tabular}{c|c}
                    AA sequence & [1, 2, 0, 3, 2, 3, 2, 0, 0, 1, 1, 3, 1, 1, 2, 0, 0, 3, 1, 1, 1, 2, 3, 0, 0, 0] \\
                    \hline
                    HA sequence &  [2, 2, -1, 3, 3, -1, 2, 2, 0, 1, 1, 3, 1, 1, 2, -1, -1, -1, 1, 3, 3, 3, 3, 0, 0, 2]\\
                    \hline
                    HA Interaction &  21  out of  26, i.e., 80.8\% \\
                    \hline
                    AA followed HA & 14\\
                    \hline
                    Reward & 20.11\\
                \end{tabular}
                \label{tab:results_case2}
        \end{table*}
        \begin{table*}[!tbh]
                \caption{Shared autonomy stats for Case 3}
                \centering
                \begin{tabular}{c|c}
                        AA sequence & [1, 2, 0, 3, 2, 3, 2, 0, 0, 1, 1, 3, 1, 1, 2, 0, 0, 3, 0, 0, 0, 2, 0, 0, 0, 3] \\
                        \hline
                        HA sequence &   [2, 2, -1, 3, 2, -1, 2, 0, 0, 1, 1, 3, 1, 1, 2, -1, -1, -1, 0, 0, 0, 2, 0, 0, 0, 3]\\
                        \hline
                        HA Interaction & 21  out of  26, i.e., 80.8\% \\
                        \hline
                        AA followed HA & 20\\
                        \hline
                        Reward & 24.26\\
                    \end{tabular}
                    \label{tab:results_case3}
            \end{table*}

        \begin{table*}[!tbh]
            \caption{Extended stats for Case 1}
            \centering
            \begin{tabular}{c|c|c|c|c|c}
                \hline
                \textbf{Test ID} & \textbf{steps} & \textbf{HA Interaction} & \textbf{AA follow} & \textbf{reward} & \textbf{success} \\ \hline
                0 & 30 & 19  & 5 & 13.96 & 1 \\ \hline
                1 & 30 & 21  & 7 & 14.92 & 1 \\ \hline
                2 & 26 & 18  & 10 & 19.46 & 1 \\ \hline
                3 & 28 & 17  & 5 & 14.21 & 1 \\ \hline
                4 & 30 & 21  & 5 & 12.53 & 1 \\ \hline
                5 & 26 & 21  & 10 & 18.18 & 1 \\ \hline
                6 & 32 & 22  & 5 & 14.44 & 1 \\ \hline
                7 & 30 & 22  & 8 & 15.70 & 1 \\ \hline
                8 & 30 & 21  & 12 & 19.14 & 1 \\ \hline
                9 & 26 & 21  & 10 & 17.20 & 1 \\ \hline \hline
                avg. & 28.8 & 20.3  & 7.7 & 15.97 & 1 \\ \hline
            \end{tabular}
            \label{tab:results_case1_extended}
        \end{table*}
        \begin{table*}[!tbh]
            \caption{Extended stats for Case 2}
            \centering
            \begin{tabular}{c|c|c|c|c|c}
                \hline
                \textbf{Test ID} & \textbf{steps} & \textbf{HA Interaction} & \textbf{AA follow} & \textbf{reward} & \textbf{success} \\ \hline
                0 & 26 & 21 & 15 & 21.30 & 1 \\ \hline
                1 & 26 & 21 & 14 & 20.08 & 1 \\ \hline
                2 & 26 & 18 & 12 & 19.13 & 1 \\ \hline
                3 & 26 & 21 & 15 & 20.18 & 1 \\ \hline
                4 & 26 & 21 & 13 & 19.08 & 1 \\ \hline
                5 & 26 & 21 & 15 & 21.06 & 1 \\ \hline
                6 & 26 & 21 & 16 & 21.20 & 1 \\ \hline
                7 & 26 & 21 & 14 & 20.70 & 1 \\ \hline
                8 & 26 & 21 & 14 & 20.56 & 1 \\ \hline
                9 & 26 & 21 & 13 & 20.05 & 1 \\ \hline \hline
                avg. & 26 & 20.7 & 14.1 & 20.34 & 1 \\ \hline
            \end{tabular}
            \label{tab:results_case2_extended}
        \end{table*}
        \begin{table*}[!tbh]
            \caption{Extended stats for Case 3}
            \centering
            \begin{tabular}{c|c|c|c|c|c}
                \hline
                \textbf{Test ID} & \textbf{steps} & \textbf{HA Interaction} & \textbf{AA follow} & \textbf{reward} & \textbf{success} \\ \hline
                0 & 26 & 21 & 20 & 24.30 & 1 \\ \hline
                1 & 26 & 21 & 18 & 22.17 & 1 \\ \hline
                2 & 26 & 21 & 19 & 23.61 & 1 \\ \hline
                3 & 26 & 21 & 19 & 23.70 & 1 \\ \hline
                4 & 26 & 21 & 17 & 22.26 & 1 \\ \hline
                5 & 26 & 21 & 19 & 23.06 & 1 \\ \hline
                6 & 26 & 21 & 19 & 23.85 & 1 \\ \hline
                7 & 26 & 21 & 18 & 23.16 & 1 \\ \hline
                8 & 26 & 21 & 16 & 20.66 & 1 \\ \hline
                9 & 26 & 21 & 19 & 23.79 & 1 \\ \hline \hline
                avg. & 26 & 21 & 18.4 & 23.06 & 1 \\ \hline
            \end{tabular}
            \label{tab:results_case3_extended}
        \end{table*}
        In this stage, we test the model trained with an expert human, discussed in Scenario I. We, however, test the model in a challenging scenario, which occurs when shared autonomy interacts with a noisy human agent that might or might not cooperate. In such scenarios, we expect a shared autonomy framework implemented with a policy shaping paradigm to outshine in its performance. For such cases, the burden of successfully carrying out the tasks is on the autonomous agent while trying as much as possible to follow the human's input. We use the model trained using raw human data, i.e., the expert human and present results for the environment initialized as shown in Figure \ref{fig:exa3_EM_fig}. We consider  the following three cases. For each case, there is range of possible outcomes due to the stochasticity of the human behavior; we present an illustrative example for each case.
        \begin{itemize}
            \item \textbf{Case 1}: Random human, i.e., very noisy or novice human. The results for this case are given in Table \ref{tab:results_case1}. In this table, the first row is the sequence of actions of autonomous agent (\textit{AA sequence}). Comparatively, the second row shows the sequence of actions of human agent (\textit{HA sequence}). The reader is reminded that an action of "-1" denotes a non-cooperative human agent. Moreover, as denoted in the third row, human agent interacted 18 times out of the total of 26 actions or $69.2\%$ of the time throughout the episode. Despite having a very noisy human, the shared autonomy managed to follow the human 8 times. This result shows that that the autonomous agent ignored the human most of time and successfully carried out the operation. Table \ref{tab:results_case1_extended} shows the extended results for 10 tests.

            \item \textbf{Case 2}: Medium-level noisy. The results for this case are given in Table \ref{tab:results_case2}, which shows how differently the autonomous agent engaged with the human compared to the previous case. This signifies the capability of the framework to discern humans with different skill levels. Table \ref{tab:results_case2_extended} shows the extended results for 10 tests.      
            
            \item \textbf{Case 3}: Least noisy human, i.e., close to expert human. The results for this case are given in Table \ref{tab:results_case2}, which once more shows how the autonomous agent engaged with the human. It is observed that the autonomous agent managed to follow this human 20 times out of 21. Table \ref{tab:results_case3_extended} shows the extended results for 10 tests.
        \end{itemize}
\section{Conclusions and Future Work} \label{sec:conclusion}
    
    Operating an articulated machine is similar to driving a car in terms of complexity and hierarchy of tasks involved, from strategical route planning to low-level controls, and it is highly intertwined with the specific requirements of the application domain. Therefore, as we argued in this paper, design for autonomous operation of such machines requires a careful understanding of the nature of the tasks and their environment. In this work, we proposed a shared autonomy framework to operate articulated robots. We first introduced a hierarchical task-oriented planning formulation for context-aware robot operation. Building on this foundation as well as theory of mind and game theory, we proposed a novel shared autonomy framework to facilitate efficient interaction between the human and the autonomy, the two participating agents in this system. We modelled the decision making process using hierarchical MDPs and Options in an algorithm we called \textit{policy shaping}. In this algorithm, the autonomous system policy is shaped by incorporating design variables contextual to the task, human's internal state, and pre-training, as well as the human's input. To encode the human's internal state beyond the designed state variables, we used the pre-trained model as the surrogate optimal model as a frame of reference to compare human's input. We employed the associated error as well as the history of states and actions in a conditional Variational Autoencoder (cVAE) architecture to find the human's latent embedding through the lens of the structured task at hand. 
    
    To showcase the success of our framework, we fine-tuned our framework for the operation of a feller-buncher articulated machine in timber harvesting, a series of physically and mentally arduous operations in harsh environmental conditions. Building on our earlier work (\cite{Yousefi2022}), with intricate know-how of the tasks, a novel, human-inspired path planning algorithm using the \textit{Envelope of Manipulation} $\mathcal{E}^M$ and the \textit{Envelope actions} to encode the sequence of decisions/actions in the operations was proposed. We have used this case study as our test-bed to train and test different policies. In training the policies, we used deep RL techniques. Moreover, by using a wide range of available tools, libraries, and packages, we setup a human-in-the-loop test that enabled us to gather actual human trials data. In presenting results, we considered a number of scenarios and cases of importance to a shared autonomy framework. First, we trained a fully autonomous policy capable of carrying out the operations in our setup. We used this model as the surrogate optimal model. By gathering actual human trials data, we were able to train a cVAE network to access a human's internal embeddings. Then, we envisioned and implemented several training scenarios involving a range of human expertise. We assessed the success of our novel platform by forming certain hypotheses regarding the affect of our designed structure and variables. In testing the trained shared autonomy policy, once more we looked at the performance of the model in interacting with human agents with different skill levels and degree of cooperativeness. The extensive test results demonstrate the success of our platform in a particularly challenging case of interacting with a noisy non-cooperative human.

    The future directions are numerous, given the potential of this novel framework. We, however, propose that the future directions should be more in-line with autonomous operation/driving scenarios, since this platform offers an alternative point of view into designing a hierarchical planning framework with full autonomy in mind. Moreover, refined training algorithms tuned for shared autonomy and human-in-the-loop scenarios as well as more structured approaches to encode human's embeddings can be considered in future work.
    

\section*{APPENDIX A} \label{app:A}
    Here, we show the derivation of \eqref{eq:dist_traj2} for 3 time-steps. For the trajectory $\tau$, we have:
    \begin{equation}
        \tau = \{s^A_1,a^A_1,a^H_1,...,s^A_3,a^A_3,a^H_3 \}, \label{eq:dist_traj_app}
    \end{equation}
    The probability over trajectory is hence given be:
    \begin{equation}
        \begin{split}
             p_{\tau} & = p(s^A_1,a^A_1,a^H_1,...,s^A_3,a^A_3,a^H_3)\\
                    & = p(a^A_1,a^H_1,...,s^A_3,a^A_3,a^H_3|s^A_1)p(s^A_1)\\
        \end{split}
    \end{equation}
    Next, we write:
    \begin{equation}
        \begin{split}
            p_{\tau} & = p(s^A_3|a^A_2, s^A_2)p(a^A_2,a^H_2,s^A_2,a^A_1,a^H_1|s^A_1)p(s^A_1)\\
                    & = p(s^A_3|a^A_2, s^A_2)p(a^A_2,a^H_2|s^A_2,a^A_1,a^H_1,s^A_1) p(s^A_2,a^A_1,a^H_1|s^A_1)p(s^A_1)\\
        \end{split}
    \end{equation}
    Next, we have:
    \begin{equation}
        \begin{split}
            p_{\tau} & = p(s^A_3|a^A_2, s^A_2)p(s^A_2|a^A_1,s^A_1)p(a^A_2|a^H_2,s^A_2) p(a^H_2|s^A_2,s^A_1)p(a^A_1|a^H_1,s^A_1)p(a^H_1|s^A_1)p(s^A_1),
        \end{split}
    \end{equation}
    which simplifies to:
    \begin{equation}\label{eq:dist_traj_simp}
        \begin{split}
            p(\tau) &= p(s^A_1) \prod_{t=1}^{3}\pi_{H}(a^H_t|\overline{s}^A_t) \pi_{A}(a^A_t|a^H_t,s^A_t)p(s_{t+1}|s^A_t,a^A_t),
        \end{split}
    \end{equation}

\section*{APPENDIX B} \label{app:noise_algorithm}
    Here we provide the pseudocode on how we perturb human action in the cases we discuss noisy human:
    \begin{algorithm}
        \caption{algorithm to perturb human input}
        \begin{algorithmic}[1]
            \State $output \gets output + \text{np.random.randint}(0,4)$
            \If{$output < 0$}
            \State $output \gets output + 4$
            \EndIf
            \If{$output > 3$}
            \State $output \gets output \mod 4$
            \EndIf
        \end{algorithmic}
    \end{algorithm}

\section*{Acknowledgement}
    We would like to thank Professor Dylan P. Losey for his early contributions to this work. 
    This work was supported by the National Sciences and Engineering Research Council (NSERC) Canadian
    Robotics Network (NCRN). The authors also acknowledge the valuable contributions of AI-Redefined and William Duguay to the development of the shared autonomy setup.

\bibliographystyle{agsm}
\bibliography{MyCollection_PhD.bib}

\end{document}